\documentclass[10pt,a4paper]{article}

\usepackage[hyperref]{acl2021}
\usepackage{times}
\usepackage{soul,color}
\usepackage{xcolor,colortbl}
\usepackage{latexsym}
\usepackage[shortlabels]{enumitem}
\usepackage{comment}
\usepackage{multirow}
\usepackage{array, makecell} 
\usepackage{graphicx}
\usepackage{subcaption}
\usepackage{multirow}
\usepackage{mathrsfs} 
\usepackage{minitoc}
\usepackage{amsmath, amssymb}
\usepackage{cleveref}
\crefname{section}{§}{§§}
\Crefname{section}{§}{§§}

\newtheorem{hypothesis}{Hypothesis}

\definecolor{Gray1}{rgb}{0.91,0.925, 0.937}
\definecolor{Gray2}{rgb}{0.87, 0.886, 0.902}
\definecolor{Gray3}{rgb}{0.808, 0.831, 0.855}
\definecolor{Gray4}{rgb}{0.678,0.71, 0.741}
\definecolor{Green0}{rgb}{0.909, 0.992, 0.886}
\definecolor{Green1}{rgb}{0.843, 0.960, 0.839}
\definecolor{Green2}{rgb}{0.635, 0.854, 0.627}
\definecolor{Red1}{rgb}{0.937, 0.686, 0.698}

\definecolor{darkgreen}{rgb}{0.0, 0.5, 0.0}
\definecolor{almond}{rgb}{0.99, 0.87, 0.9}
\definecolor{ghostwhite}{rgb}{0.97, 0.97, 1.0}
\definecolor{Blue1}{rgb}{0.792, 0.941, 0.973}
\definecolor{Blue2}{rgb}{0.678, 0.91, 0.957}
\definecolor{Blue3}{rgb}{0.565, 0.878, 0.937}
\definecolor{Blue4}{rgb}{0.282, 0.749, 0.89}

\definecolor{Yellow1}{rgb}{1, 0.914, 0.306}
\definecolor{Yellow2}{rgb}{1, 0.886, 0.275}
\definecolor{Yellow3}{rgb}{1, 0.855, 0.239}

\usepackage{microtype}

\aclfinalcopy 

\title{
Zero-Shot Controlled
Generation with Encoder-Decoder Transformers}

\author{Devamanyu Hazarika \\
    Amazon Alexa AI\\
  \texttt{dvhaz@amazon.com} \\\And
  Mahdi Namazifar\thanks{\hspace{2mm}Corresponding author} \\
  Amazon Alexa AI\\
  \texttt{mahdinam@amazon.com} \\\And
  Dilek Hakkani-Tür \\
  Amazon Alexa AI\\
  \texttt{hakkanit@amazon.com} \\}

\date{}

\begin{document}
\maketitle
\begin{abstract}
Controlling neural network-based models for natural language generation (NLG) has broad applications in numerous areas such as machine translation, document summarization, and dialog systems. Approaches that enable such control in a zero-shot manner would be of great importance as, among other reasons, they remove the need for additional annotated data and training.
In this work, we propose novel approaches for controlling encoder-decoder transformer-based NLG models in zero-shot. This is done by introducing three control knobs, namely, attention biasing, decoder mixing, and context augmentation, that are applied to these models at generation time. These knobs control the generation process by directly manipulating trained NLG models (e.g., biasing cross-attention layers) to realize the desired attributes in the generated outputs. We show that not only are these NLG models robust to such manipulations, but also their behavior could be controlled without an impact on their generation performance. These results, to the best of our knowledge, are the first of their kind. Through these control knobs, we also investigate the role of transformer decoder's self-attention module and show strong evidence that its primary role is maintaining fluency of sentences generated by these models. Based on this hypothesis, we show that alternative architectures for transformer decoders could be viable options. We also study how this hypothesis could lead to more efficient ways for training encoder-decoder transformer models.
\end{abstract}

\section{Introduction}
\label{sec:intro}

Natural language generation (NLG) aims at producing fluent and coherent sentences and phrases in different problem settings such as dialog systems~\cite{DBLP:journals/tois/HuangZG20}, machine translation~\cite{DBLP:journals/corr/abs-2002-07526}, text summarization~\cite{DBLP:journals/access/SyedGM21}. Due to their outstanding power in recognizing patterns and generalization, neural network-based models have dominated NLG research in the past decade. Most recently, the majority of the research in NLG leverages transformers \cite{NIPS2017_3f5ee243} and specifically, transformer decoders to generate natural language~\cite{radford2019language,brown2020language,DBLP:conf/acl/LewisLGGMLSZ20}. As a general paradigm, in these approaches, natural language is generated autoregressively one token at a time, and each token is generated based on an inferred probability distribution over all possible tokens. Although these statistical approaches to NLG have proven to be highly effective, their stochastic nature and complex architectures make them difficult to control in order for them to reflect any set of desired attributes in the output. These attributes could range from persona, sentiment, condolence, dialog acts, questions, for dialog response generation~\cite{DBLP:journals/tacl/NiuB18,DBLP:conf/acl/KielaWZDUS18,DBLP:conf/naacl/SeeRKW19,DBLP:journals/corr/abs-2008-12579} to story ending control for story generation~\cite{peng-etal-2018-towards} or formality and politeness control for drafting emails~\cite{DBLP:conf/acl/MadaanSPPNYSBP20}, amongst others. 

\begin{figure}[t!]
    \centering
    \includegraphics[width=0.99\linewidth]{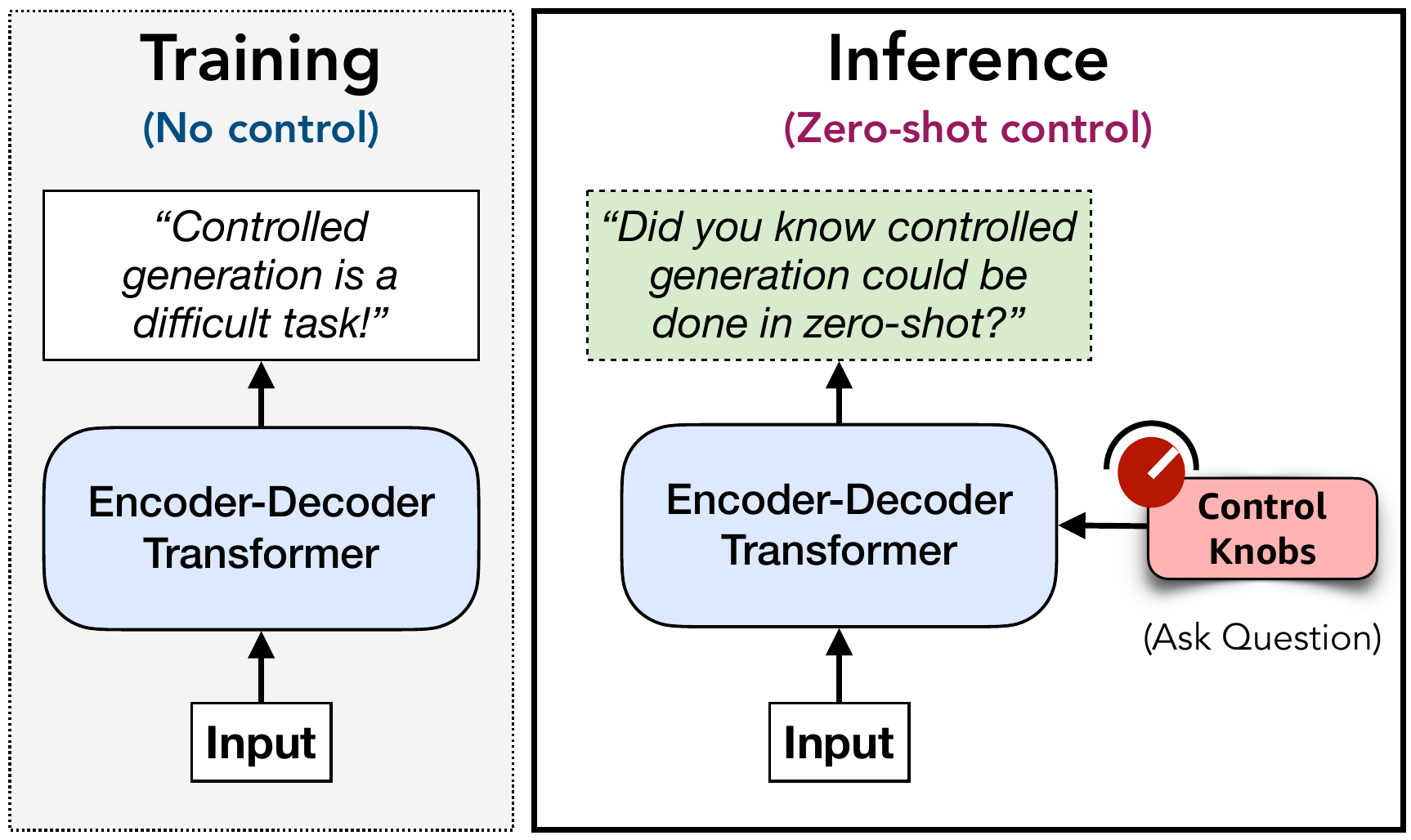}
    \caption{\footnotesize Performing zero-shot controlled generation on an encoder-decoder transformer at inference time using control knobs. The control knobs influence the generation process in such a way that the generated output has the desired attributes (e.g., asking questions).}
    \label{fig:zero-shot-control}
\end{figure}

\begin{figure*}[t!]
    \centering
    \includegraphics[width=.9\textwidth]{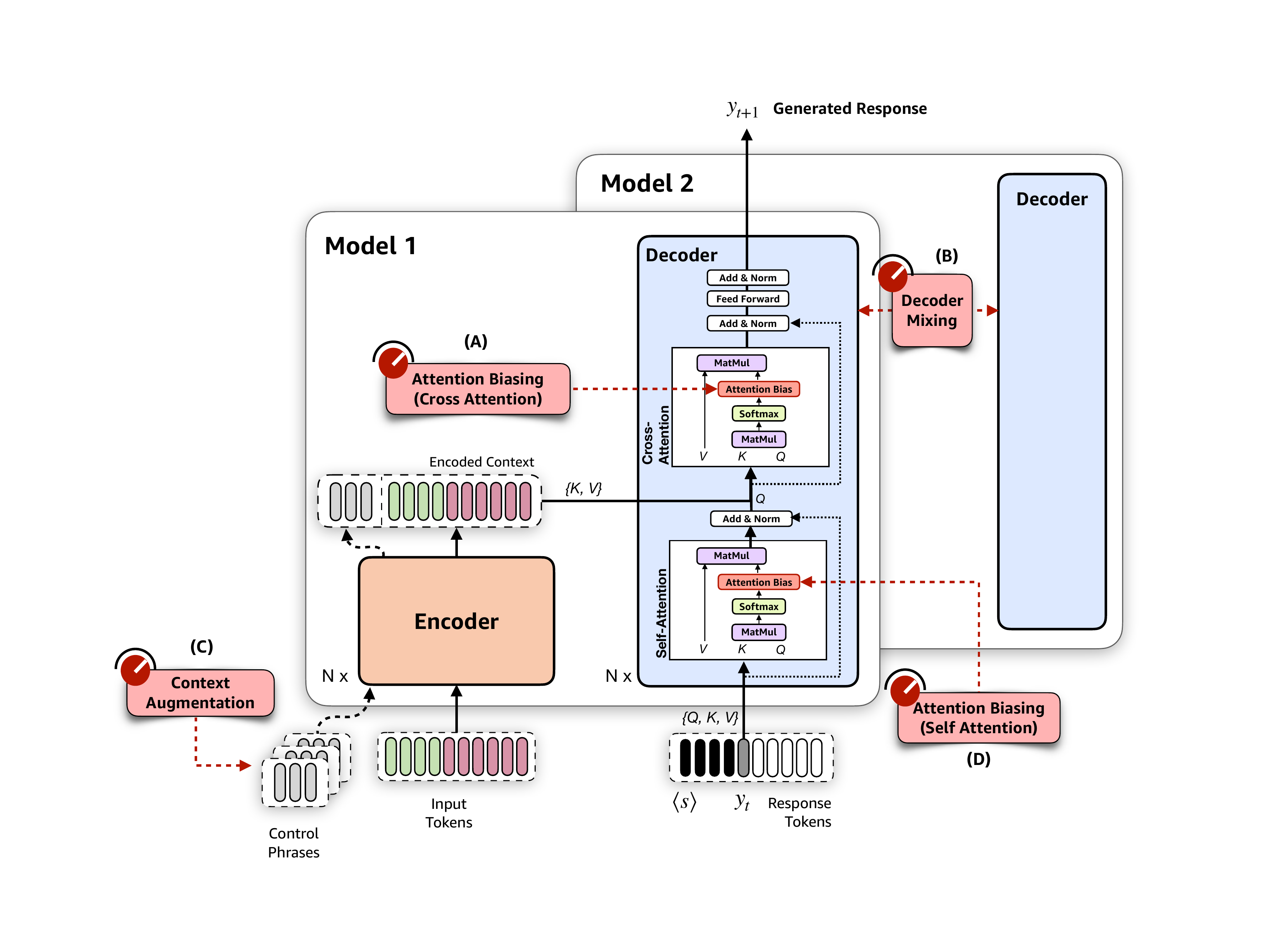}
    \caption{\footnotesize Control knobs for zero-shot controlled NLG: (A) attention biasing knob for cross-attention, (B) decoder mixing knob, (C) context augmentation knob, and (D) attention biasing knob for self-attention}
    \label{fig:overall}
\end{figure*}

In general, being able to control an NLG model in a zero-shot fashion would be highly instrumental since such zero-shot control would not require
large amounts of annotated data, nor would it require any fine-tuning of parameters of the NLG model or auxiliary attribute models to guide the generation. In this work, we introduce new zero-shot approaches for controlling NLG models based on encoder-decoder transformer architectures (that in the interest of brevity we refer to as EDT-NLG). 
The high-level idea of these approaches is to explicitly manipulate the transformers within the trained EDT-NLG models to achieve the desired attributes at generation time (see~\Cref{fig:zero-shot-control}). More specifically, we introduce a set of three \textit{control knobs}, namely, \textit{attention biasing},  \textit{decoder mixing} and \textit{context augmentation}
that could be used to control the generation of EDT-NLG models. 
These knobs could be modulated to achieve varying degrees of control in generation.
The attention biasing knob modulates the amount of attention paid to different parts of the attention context (i.e., what the attention module attends to). The decoder mixing knob works based on the idea that different decoder transformers with different learned behaviors such as input reconstruction (in auto-encoders), summarization, response generation, etc., could be combined at generation time to achieve mix of these behaviors in the generation process. The context augmentation knob works based on introducing additional context on the encoder side to generate as per desired attributes. Here, we note that there is no control-specific training in these approaches and no gradient update is involved in applying these knobs. An overview of how these control knobs function is shown in \Cref{fig:overall}. In the next sections, we will describe in detail how these knobs work, and through computational results, we show that these knobs are highly effective in zero-shot controlling of the generation in EDT-NLG models.

With that in mind, it is quite unexpected and counter-intuitive that according to the experiments results (shown in \Cref{sec:informativeness_experiments} and \Cref{sec:control_experiments}), manipulation of trained attention layers and transformers in general, through control knobs, does not derail EDT-NLG models. This robustness of EDT-NLG models to the manipulations introduced through control knobs raises many new questions. One of such questions is about the limits of such manipulations and when these manipulations cause the models to break down. Additionally, what are some of the implications of the robustness of these models to such manipulations? We address some of these questions in \Cref{sec:analysis}, where we show strong evidence that in EDT-NLG models fluency of the generation is managed by the decoder self-attention. Based on these results, we investigate alternative architectures for transformer decoders, as well as approaches for more efficient training of EDT-NLG models.

To summarize, this work's contributions are as follows:
\begin{itemize}[leftmargin=*]
    \item We propose a set of control knobs that can control EDT-NLG models during generation in a zero-shot manner, i.e., without training for controlled generation or using any gradient-based optimization during inference.
    \item We explore the application of the proposed control knobs for knowledge-grounded response generation and find that these control knobs can achieve zero-shot controlled generations, for a wide variety of attributes.
    \item We put forth and analyze the hypothesis that in EDT-NLG models fluency of generation is managed by decoder self-attention. Based on this analysis, we also explore alternative architectures for transformer decoders and propose efficient ways of training EDT-NLG models.
\end{itemize}

The control knobs introduced in this paper could be generalized to any EDT-NLG model for any NLG task. Moreover, the attention biasing knob is generic to any attention mechanism within or outside of a transformer-based architecture and could be applied to other attention-based applications such as computer vision and multi-modal problems. However, to demonstrate the efficacy of the control knobs, in this work we focus on a specific family of NLG tasks, namely knowledge-grounded open-domain Neural Response Generation (K-NRG). We use K-NRG to present the ideas,  experiments, and computational results. 

\section{Related Works}
Numerous works in the literature focus on controlling neural network-based NLG models~\cite{DBLP:conf/coling/PrabhumoyeBS20}. These approaches fall under two major categories. The first category focuses on using data annotated with the desired attributes to train the NLG model such that it is able to generate with the same attributes \cite{DBLP:journals/corr/abs-1909-05858,DBLP:journals/corr/abs-2005-00613,DBLP:journals/corr/abs-2009-10855,DBLP:conf/naacl/SeeRKW19}. The drawback of this approach is that for every set of desired attributes, annotated training datasets are required, which makes this approach difficult to scale. It also makes it difficult for trained models to generalize to other forms of control.  Furthermore, often times, there is a different dataset per desired attribute, and it is not clear how to combine multiple attributes as they may override the effect of each other.

The second category involves approaches that do not alter the model parameters but rather modify the decoding strategies during inference. To achieve the desired control, these approaches \textit{nudge} or \textit{reweigh} the output distribution towards the desired directions, either using discriminators~\cite{DBLP:conf/acl/ChoiBGHBF18} or bag-of-words that are indicative of the attributes~\cite{DBLP:conf/acl/GhazvininejadSP17,DBLP:conf/emnlp/BahetiRLD18,DBLP:conf/naacl/SeeRKW19}. However, these kinds of decoding strategies have been observed to be brittle, particularly for tasks like dialog response generation~\cite{DBLP:conf/naacl/SeeRKW19}. Another set of approaches within this category leverage auxiliary models that can detect the desired attributes. Termed as Plug-and-Play Language Models (PPLM) \cite{DBLP:journals/corr/abs-1912-02164,DBLP:conf/emnlp/MadottoILDF20}, these approaches utilize the gradients from the auxiliary attribute detection model, using which the generative model performs optimization and increases the likelihood of receiving a high score from the auxiliary model. In these methods training auxiliary models still require annotated data that could be expensive to acquire. Moreover, approaches like PPLM that employ gradient updates at generation time are computationally expensive to generate. Note that while in the first category of controlled NLG approaches, gradient updates are used at training time, in the second category, gradient updates are rather indirect and occur at generation time only. 

In contrast to the above categories, the goal of this work is controlled NLG in zero-shot. In the recent years, in machine learning in general, there has been an increased emphasis on zero-shot approaches that do not require any specific gradient-based optimization (neither during training nor during inference). In particular, prompt-based approaches have been proposed that prime massive language models, like GPT-3~\cite{brown2020language}, with few-shot supervised examples of a specific task. These approaches have led to favorable results  where the model is able to adapt to new tasks without any fine-tuning. However, to the best of our knowledge, there is no work in the literature focused on controlling the output within an NLG task in a zero-shot setting~\cite{DBLP:journals/corr/abs-2005-00613} through priming of language models. And also, to the best of our knowledge, this work is the first to propose approaches for zero-shot controlled NLG.

\begin{figure}[t!]
    \centering
    \includegraphics[width=0.9\linewidth]{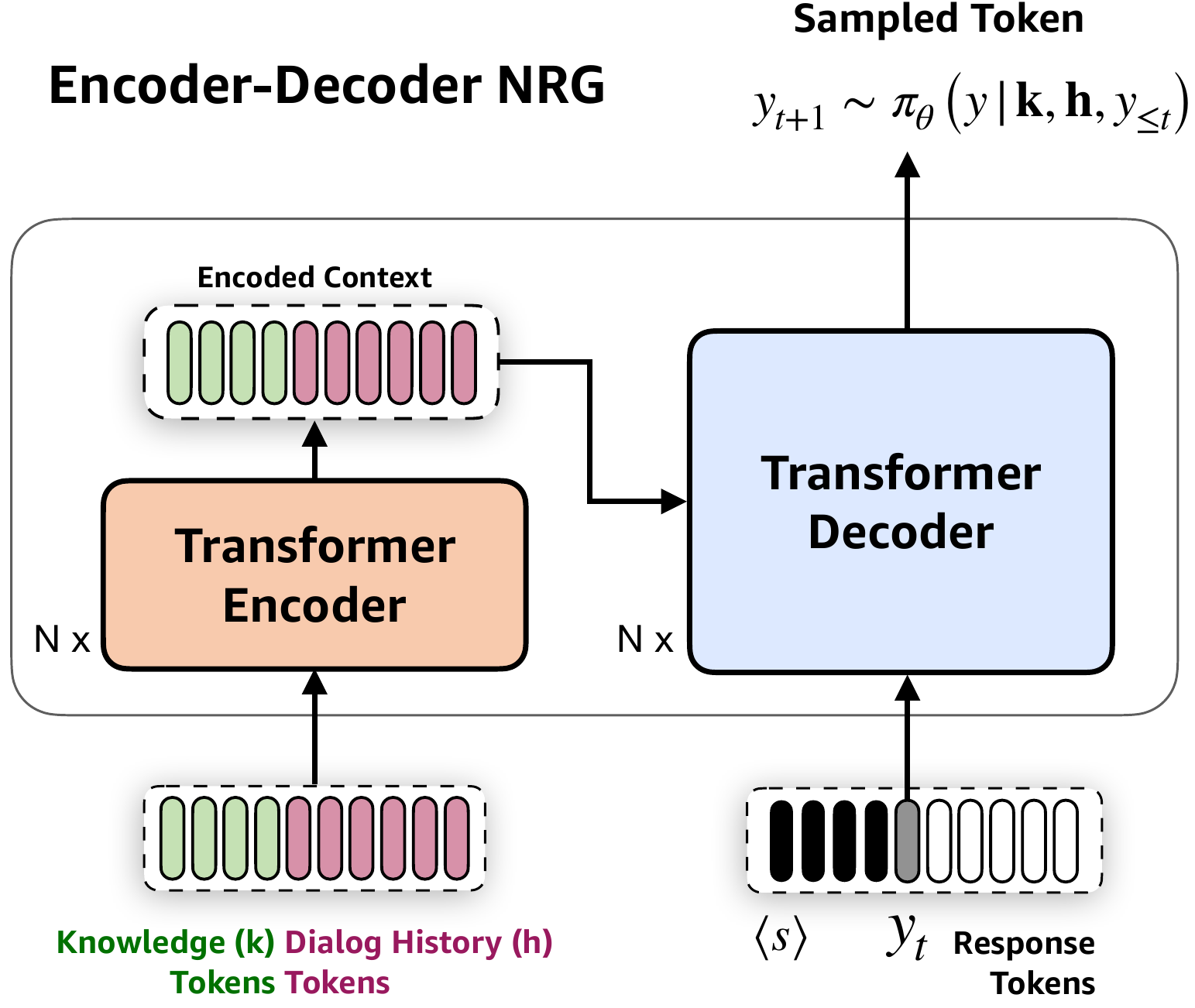}
    \caption{\footnotesize Overall architecture of encoder-decoder transformer-based model for K-NRG problem.}
    \label{fig:seq_trans}
\end{figure}

\section{Control Knobs for Zero-Shot Controlled NLG}
\label{sec:knobs}

This section discusses the details of control knobs, namely attention biasing, decoder mixing, and context augmentation, that are proposed for zero-shot controlled NLG. These knobs are intuitively designed such that the weights and outputs of attention layers in a trained NLG model are modified at inference time to achieve desired attributes in the generated outputs. 

\begin{figure}[t!]
   \centering
   \includegraphics[width=0.7\linewidth]{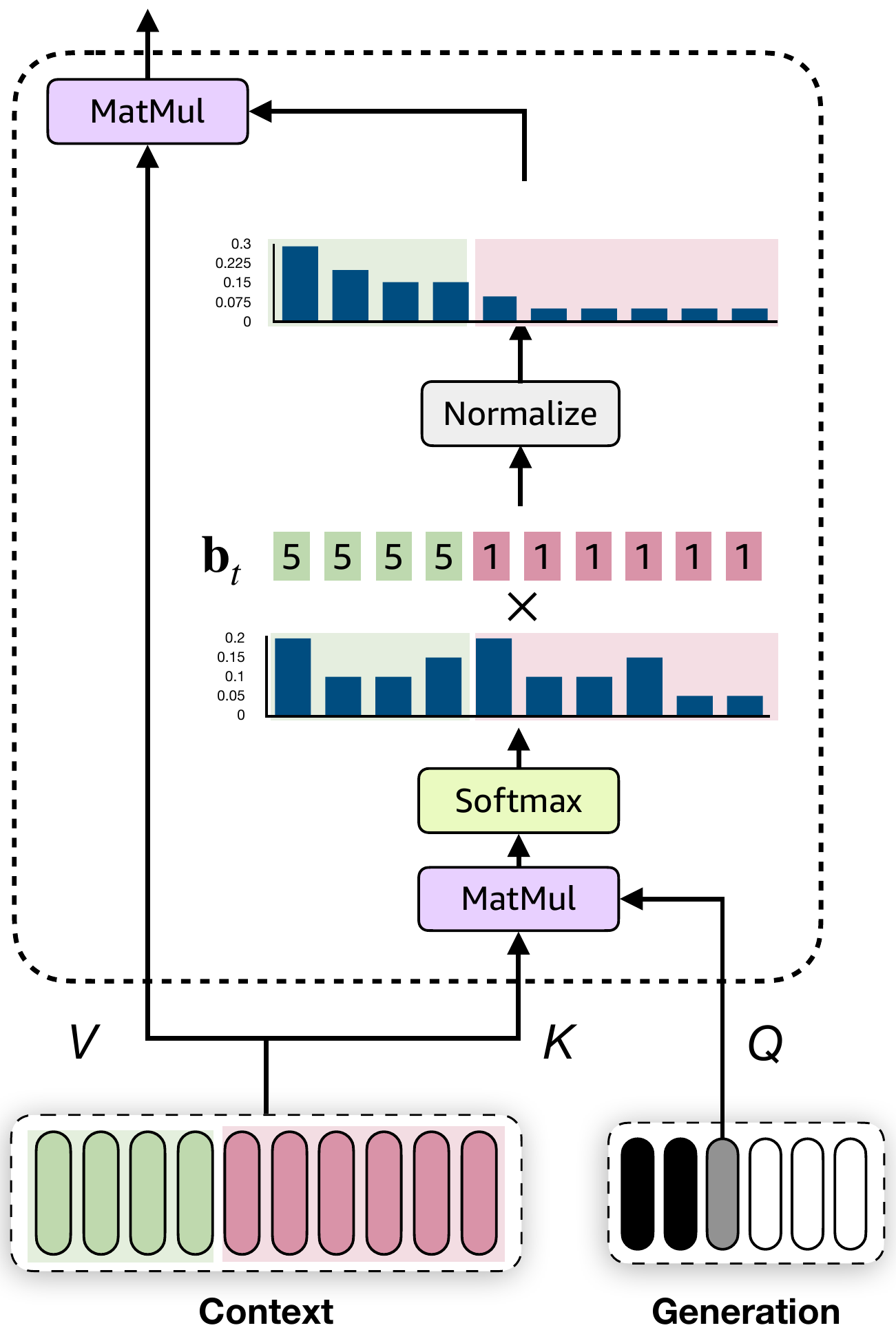}
   \caption{ \footnotesize Details of the attention biasing knob for cross-attention. Values 1 and 5 are example attention bias values.}
   \label{fig:xatt-&-xatt-bias}
\end{figure}

As a high-level summary of how the control knobs for zero-shot controlled NLG work, consider a trained EDT-NLG model $\pi_\theta$, where $\theta$ represents the parameters of the model, and $\pi_\theta$ is conditioned on a context $x$. The generation process using this model could be represented as sampling a response $y$ from $\pi_\theta$, i.e., $y \sim \pi_\theta(y|x)$. Generating with additional desired attributes, e.g., positive sentiment, could be interpreted as introducing an additional condition $c$ to the sampling process. The control knobs manually modify $\pi_\theta$ to $\Tilde{\pi}_{\Tilde{\theta}}$ such that samples from $\Tilde{\pi}_{\tilde{\theta}}(y|x,c)$ on average manifest the desired attributes significantly more.

\subsection{Attention Biasing}
\label{sec:attention_biasing}

We describe how the \textit{attention biasing} knob works for cross-attention in EDT models. Consider the cross-attention layer in the decoder as shown in \Cref{fig:xatt-&-xatt-bias}. At generation time step $t$, the decoder attends to the context in the following manner: the query vector is first multiplied by the key matrix, and the result goes through a Softmax operation that outputs a discrete probability distribution, also referred to as \textit{attention distribution}. Attention distribution is then used (through multiplication with the value matrix) to determine in some sense how much attention should be paid to each one of the context tokens \cite{DBLP:conf/iclr/DanilukRW017}. 

The idea of the attention biasing knob is forcing an attention module to attend to some parts of its context more or less than it usually would, by directly adjusting the attention distribution. We do this through element-wise multiplication of a \textit{bias vector} with the attention distribution and then normalizing the results so that the outcome is still a probability distribution (referred to as biased attention distribution). As an example, in \Cref{fig:xatt-&-xatt-bias} the cross-attention context has two parts, and the attention process is biased by multiplying the attention to the first part by some value (for example, 5) and then normalizing the outcome to retrieve a probability distribution.

More formally, given embedded context $C$, attention matrices $W_K$, $W_V$, $W_Q$, and the embedding $\mathbf{e}_t \in \mathbb{R}^d$ for $y_t$, cross-attention output for $y_t$ is: 
\begin{equation*}
    \text{softmax}\left(\frac{(\mathbf{e}_t W_Q)(CW_K)^T}{\sqrt{d}}\right)CW_V
\end{equation*}
In this notation, biased cross-attention could be defined as:
\begin{equation}
    \label{eq:att-bias}
    \mathcal{N}\left(\mathbf{b}_t \odot \text{softmax}\left(\frac{(\mathbf{e}_t W_Q)(CW_K)^T}{\sqrt{d}}\right)\right)CW_V,
\end{equation}
where function $\mathcal{N}$ normalizes a given positive vector to have the element-wise sum of 1, $\mathbf{b}_t$ is the bias vector at time step $t$, and $\odot$ represents element-wise vector multiplication. 

Attention biasing for self-attention works similarly to its cross-attention counterpart, which we discuss at length in \Cref{sec:analysis}. Note that in this work, vector $\mathbf{b}_t$ is not a learned parameter, and it is manually set so that, similar to probing, the effects of intuitive designs for this vector could be analyzed. 

Biasing of attention modules has been employed in applications such as machine translation, to achieve local or focused attention. These include learning local windows of attention using strategies like gaussian-based biases~\cite{luong2015effective,yang2018modeling}, hard-coded biases~\cite{you2020hard}, etc. Attention biases could also be induced using relative embeddings~\cite{shaw2018self}. Another popular way to bias attention is by learning differentiable masks~\cite{nguyen2020differentiable,fan2021mask}. Similar to these works, the attention biasing knob also biases the attention distribution, but unlike these works, in the attention biasing knob the  bias is applied in zero-shot and on a continuous scale. While zero-shot biases have been studied in the probing literature to understand the influence of attentions on model's classifications~\cite{serrano2019attention}, to our knowledge, zero-shot attention biasing for controlled generation is an unexplored avenue.

\begin{figure}[t!]
    \centering
    \includegraphics[width=0.95\linewidth]{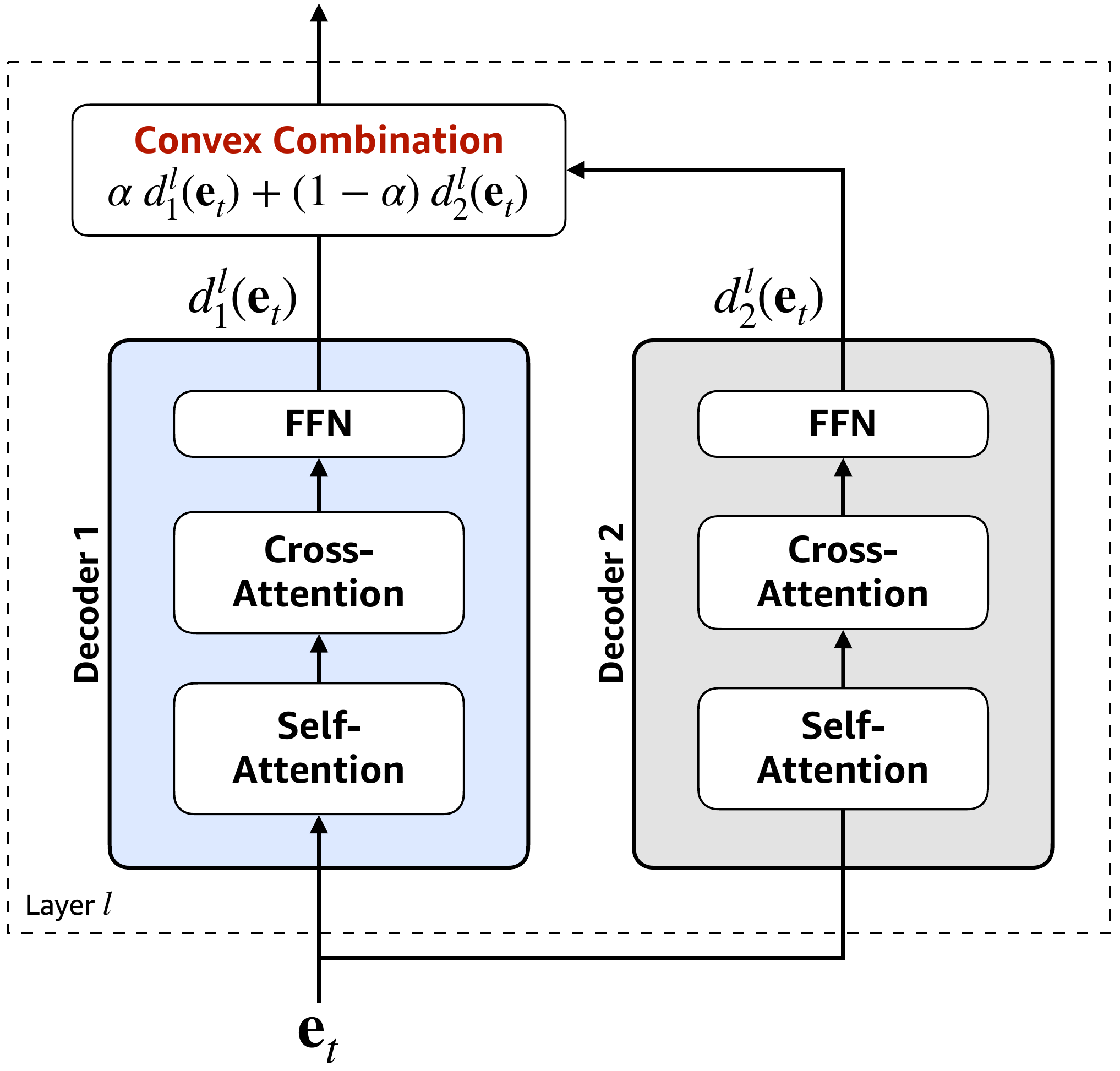}
    \caption{\footnotesize Example of decoder mixing with two decoders. For every decoding time step and decoder layer, a convex combination is applied to the output of the two decoder blocks.}
    \label{fig:decoder-mix}
\end{figure}

\subsection{Decoder Mixing}
\label{sec:decoder_mixing}

The \textit{decoder mixing} knob, as the name suggests, mixes two or more trained transformer decoders at every layer. Inspired by cross-stitch networks for multi-task learning~\cite{DBLP:conf/cvpr/MisraSGH16}, the mixing in this knob is done through convex combination of the output of the transformer decoders for every decoder layer. Figure \ref{fig:decoder-mix} shows how decoder mixing works for two decoders. 

The intuition behind this control knob is combining different behaviors that are learned in different decoders. For instance, consider a BART model in which the decoder has learned to reconstruct the input to the model and, and as a result, could be thought of as having a copying (of the context) behavior. Also consider another EDT-NLG model in which the decoder has learned to generate a response given a dialog, hence it has a responding behavior. In this example, the decoder mixing knob could be used for combining these two behaviors to generate responses where parts of the context (e.g., knowledge) is incorporated (more or less copied) in the generated response. 


Combining multiple decoders has not been extensively explored in the literature. One notable work is \cite{DBLP:journals/tacl/NiuB18}, where the authors propose late fusion of two decoders by merging the probability distribution predicted from each decoder. 
The main difference between the late fusion approach and the decoder mixing knob is that in this knob the mixing is done at every layer of the decoder and not at the output layer. 

In the decoder mixing knob, at every generation time step $t$ and decoder layer $l$, two or more transformer decoders are applied followed by a convex combination of the output of these decoders. More formally, if there are $n$ transformer decoders and transformer decoder $i$ is represented as function $d_i^l$ for all $i = 1, ..., n$, the output of decoding at time step $t$ for the input $\mathbf{e}_t$ is equal to $\sum_i \alpha_i d_i^l(\mathbf{e}_t)$ where $\alpha_i \in [0,1], \forall i$ and $\sum_i \alpha_i=1$. This mixing process is repeated across all the layers of the decoders. We refer to the vector of $\alpha$ values as the \textit{decoder mixing vector} and represent it as $\boldsymbol{\alpha}$.

\subsection{Context Augmentation} 
\label{sec:context_biasing_knob}

\begin{figure}[t!]
    \centering
    \includegraphics[width=\linewidth]{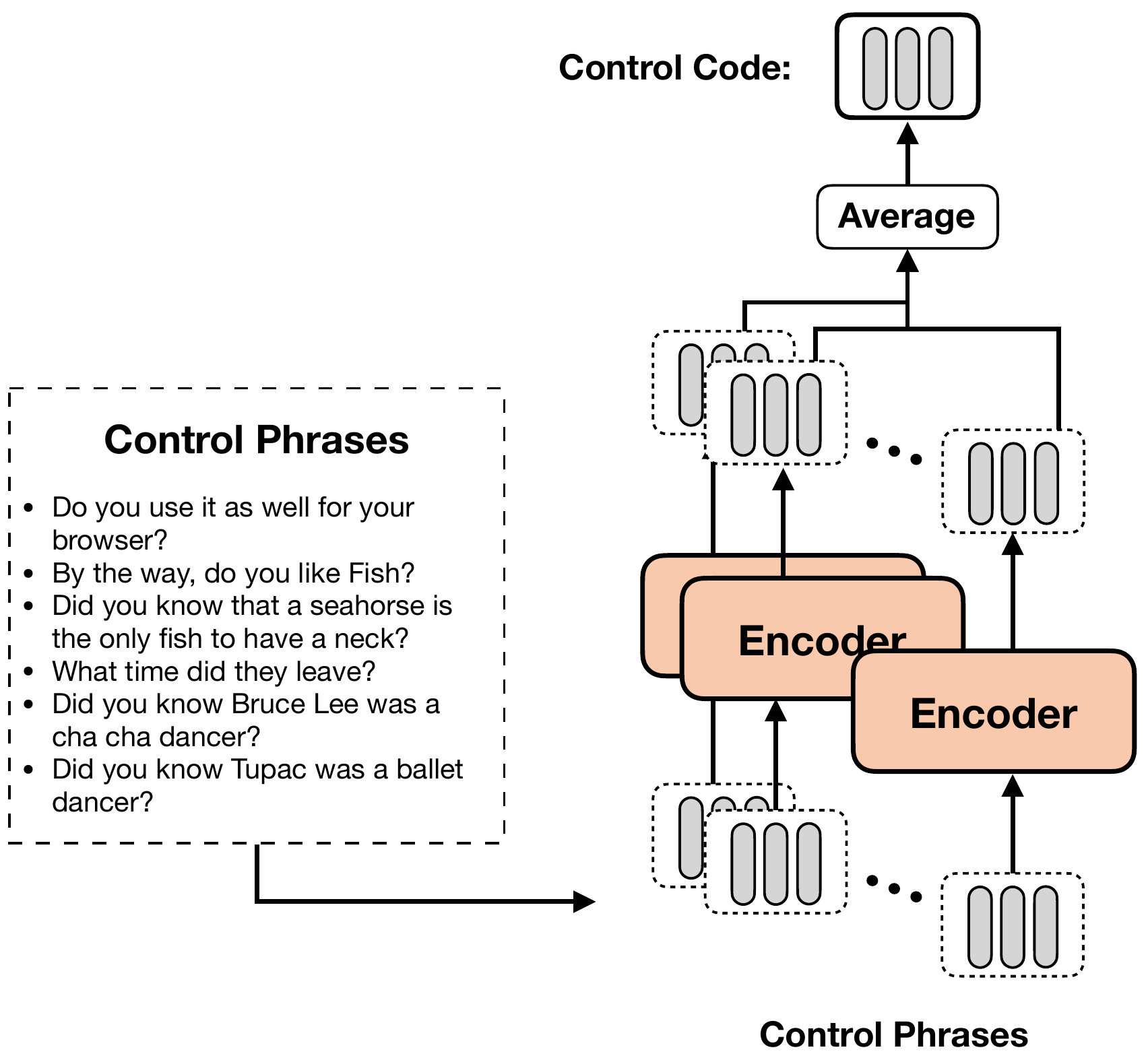}
    \caption{\footnotesize Example of creating control codes for questions.}
    \label{fig:biasing-code}
\end{figure}

In the \textit{context augmentation} knob, we apply modifications to the input of the EDT-NLG model in order to push the model to manifest the desired attributes in the generations. 
We explain how the context augmentation knob works through an example. Imagine that the desired attribute for the output of the model is \textit{asking a question}, i.e., inquisitive generation. In other words, we would like to increase the likelihood of the model's output including a question. To this end, we first sample a set of question sentences (e.g., by choosing sentences that end with a question mark) from any text corpora. We call these sentences \textit{control phrases}. We then feed each control phrase to the encoder of the EDT-NLG model to get an embedding for it\footnote{The embedding of a particular control phrase is the contextual sequence of token representations generated by the encoder of the encoder-decoder model.}. We then take the average of these embeddings across all control phrases (Figure \ref{fig:biasing-code}). We refer to this average as \textit{control code}. The control code is then concatenated to the encoded context as shown in part (C) of Figure \ref{fig:overall}.

The control code, being the average of embedded control phrases, is designed to capture the shared concepts among the control phrases. The role of averaging in creating control codes is to maintain the shared concepts within the control phrases (which is question in our example) and smoothing out other concepts such as topic. This concept of an average control code is inspired from the \textit{prototypes} in~\cite{DBLP:conf/nips/SnellSZ17}. 


\section{Overview of Experiments}
\label{sec:exp}
We conduct extensive experiments and ablations to understand the efficacy of the control knobs introduced in the previous section. This section presents an overview of the experiments and establishes the different settings under which they are conducted.

\subsection{Preliminaries: Knowledge-Grounded Open-Domain NRG}
\label{sec:exp-kgnrg}

\begin{table}[t!]
\centering\renewcommand\cellalign{lc}
\setcellgapes{3pt}\makegapedcells
\resizebox{\linewidth}{!}{
    \begin{tabular}{|r|l|} \hline
        \multirow{3}{*}{\makecell{\textbf{Previous} \\ \textbf{Turns}}} & \textbf{A}: Hi!  do you like to dance?\\
        & \makecell{\textbf{B}: I love to dance a lot. How about you?}\\
        & \makecell{\textbf{A}: I am really bad, but it is a good time.}\\
        \hline
        \makecell{\textbf{Knowledge}} & \makecell{\itshape ``Bruce Lee was also a great dancer and \\ \itshape that he won the Hong Kong Cha-Cha \itshape  \\ \itshape Championship in 1958.''} \\ 
        \hline
        \makecell{\textbf{Next Turn}} & \\
        \colorbox{ghostwhite}{uninformative} & \makecell{\textbf{B}: Hmm. Dancing is a lot of fun.}\\
        \colorbox{Blue1}{informative}&\makecell{\textbf{B}: Dancing is a lot of fun. \underline{Even Bruce Lee} \\ \underline{was a great dancer and has won competitions.}}\\
        \colorbox{Green1}{inquisitive}&\makecell{\textbf{B}: Dancing is a lot of fun. \underline{Did you know} \\ \underline{that Bruce Lee was a great dancer?}}\\
        \hline
    \end{tabular}
}
\caption{ \footnotesize An example of knowledge-grounded response generation from Topical-Chat dataset \cite{Gopalakrishnan2019}. For the provided conversational context, multiple styles of responses (such as more informative or inquisitive) are appropriate.}
\label{tab:tcs_ex}
\end{table}

We use the K-NRG problem in our experiments to study the efficacy of the control knobs for controlled generation.  In this problem, the input comprises the previous dialog turns (also referred to as dialog history) and one or more knowledge snippets related to the dialog. The task is to generate the next turn of the dialog. \Cref{tab:tcs_ex} shows an example of the previous turns of a dialog and a provided knowledge snippet, as well as the successive turn of the dialog.

For the K-NRG problem, we train encoder-decoder transformer-based NRG (EDT-NRG) models.
More specifically, at every turn, the previous dialog turns $h$ is concatenated to the provided knowledge snippet $k$, and the result $(k,h)$, collectively referred to as dialog context, is encoded by the encoder (see~\Cref{sec:appendix_input_formatting}). The decoder is prompted with the start of the sentence token $\big \langle s \big\rangle$ and in an auto-regressive manner generates one token ($y_{t+1}$) at a time, based on cross-attention to the encoded dialog context and self-attention to the previously generated tokens ($y_1, y_2, ..., y_t$) until a special end-of-sentence token is generated, i.e.
\begin{align*}
y_{t+1}\sim \pi_{\theta}(y|\mathbf{k},\mathbf{h},y_1, ..., y_t),
\end{align*}
where $\pi_\theta$ is the EDT-NRG model with parameters $\theta$. 


We use the setting introduced in the Topical-Chat dataset~\cite{Gopalakrishnan2019} which includes dialogues between two Mechanical Turk workers (a.k.a. Turkers). 
Based on the previous work \cite{DBLP:conf/inlg/HedayatniaGKLEH20}, we choose the setting where for each turn in the dialog, the knowledge snippet that is the most similar to the ground truth response is selected using 
TF-IDF and is provided as additional input.


For the NRG model, we use BART as the pre-trained encoder-decoder transformer model (EDT)~\cite{DBLP:conf/acl/LewisLGGMLSZ20}. In particular, we choose the smaller BART-base model for two reasons. First, smaller models require significantly less compute resources and are more economical with a much less carbon footprint. Second, they are more challenging for zero-shot control as previous results highlight the difficulty to achieve zero or few-shot capabilities in smaller models~\cite{DBLP:journals/corr/abs-2009-07118}. Full details over the fine-tuning procedure of the Bart-base model on the Topical-Chat dataset is provided in~\Cref{sec:appendix_model_details}.

We evaluate the efficacy of the control knobs over the two \textit{frequent} and \textit{rare} test sets from the Topical-Chat dataset. As the name suggests, the frequent test set contains entities in the dialogs that frequently appear in the training set, whereas the rare test set contains entities that are not frequent in the training data.

\subsection{Goals of the Experiments} 
\label{sec:experimental_goals}

The goals of experiments in this work are two-fold. First, we examine whether the proposed control knobs effectively control the generation process to generate according to desired attributes. Second, we examine whether applying the knobs would cause negative impacts on the generation output. Specifically, we examine the impact of the control knobs on \textit{fluency} and \textit{relevance} of the generated response. Fluency refers to the grammatical and syntactical correctness of generated responses. Relevance refers to appropriateness of a response given the history of the dialog~\cite{DBLP:conf/naacl/SeeRKW19,DBLP:journals/corr/abs-1906-08487,DBLP:conf/acl/RashkinSLB19}.

It should be re-emphasized that our primary goal is to explore the zero-shot controllability of EDT-NLG models using the control knobs. As such, we do not intend to propose a general-purpose response generator that leverages such controllable generations to improve the overall conversation experience. Such models would need appropriate dialog policies, such as dialog act-based policies~\cite{DBLP:conf/inlg/HedayatniaGKLEH20,DBLP:conf/sigdial/SankarR19} or other content planners~\cite{DBLP:journals/corr/abs-2005-00613}, and we leave the exploration of these as future work.

Due to the differences between the attention biasing and decoder mixing knobs on the one hand, and the context augmentation knob on the other hand, we split the experiments into two sections. In \Cref{sec:informativeness_experiments} we present the experiments for cross-attention biasing and decoder mixing knobs. In \cref{sec:control_experiments} we discuss the experiments for the context augmentation knob. After the experiment sections, we discuss in details self-attention biasing, alternative architectures for transformer decoders, and more efficient training of EDT-NLG models in \Cref{sec:analysis}. All of our experiments are done across five runs to account for variability in the token sampling procedure.

\section{Experiments: Attention Biasing and Decoder Mixing}
\label{sec:informativeness_experiments}

\subsection{Experiments Setup}
\label{sec:informativeness_bias_profiles}

In this section, we study the effects of applying cross-attention biasing (\Cref{sec:attention_biasing}) and decoder mixing knobs (\Cref{sec:decoder_mixing}) for zero-shot controlled NLG. We focus our experiments on controlling informativeness in generated responses for the Topical-Chat problem setup introduced in \Cref{sec:exp-kgnrg}. 
As a reminder, in this setup, input $x$ is composed of dialog history $h$ and a knowledge snippet $k$, i.e., $x = (k, h)$. 



\subsubsection{Cross-Attention Biasing Profiles}
\label{sec:xatt-profiles}

We first apply the attention biasing knob for cross-attention modules on the transformer decoder of an EDT-NRG model fine-tuned for Topical-Chat. As the dialog context is a sequence with two parts, knowledge snippet $k$ and dialog history $h$ (\Cref{fig:xatt-&-xatt-bias}), the bias vector at generation time step $t$ could be represented as the row vector $\mathbf{b}_{t}$ which is the concatenation of two bias row vectors $\mathbf{b}^k_{t}$ and $\mathbf{b}^h_{t}$, i.e., $\mathbf{b}_{t} = [\mathbf{b}^k_{t};\mathbf{b}^h_{t}]$. Although these vectors can be composed of different elements, meaning that at time $t$ the attention bias value for $i^{\rm th}$ token of knowledge snippet could be different from that of $j^{\rm th}$ token, we simplify the setup by assigning one attention bias value for knowledge ($b^k_t$) and another for dialog history ($b^h_t$) for each generation time step $t$. In other words:
\begin{align*}
    \mathbf{b}_{t} &= [\mathbf{b}^k_{t};\mathbf{b}^h_{t}] \\
    \mathbf{b}^k_{t} &= \left[\left(b^k_{t}\right)_{\times |k|}\right] \quad \text{and} \quad 
    \mathbf{b}^h_{t} &=  \left[\left(b^h_{t}\right)_{\times |h|}\right], 
\end{align*}
where, $|k|$ and $|h|$ represent the total number of tokens in knowledge and dialog history, respectively. As a hypothetical example if  $k$ has 3 tokens and $h$ has 4 tokens, and at time step $t$ we give attention bias value of 5 to knowledge ($b_t^k = 5$) and 1 to dialog history ($b_t^h = 1$), then $\mathbf{b}_t = [5,5,5,1,1,1,1]$.

Following this notation, we design different biasing profiles to explore the extent of controlled generations we can achieve from biasing cross-attention through the attention biasing knob. We experiment with three different biasing profiles, namely:

\begin{figure}[t!]
    \centering
    \includegraphics[width=\linewidth]{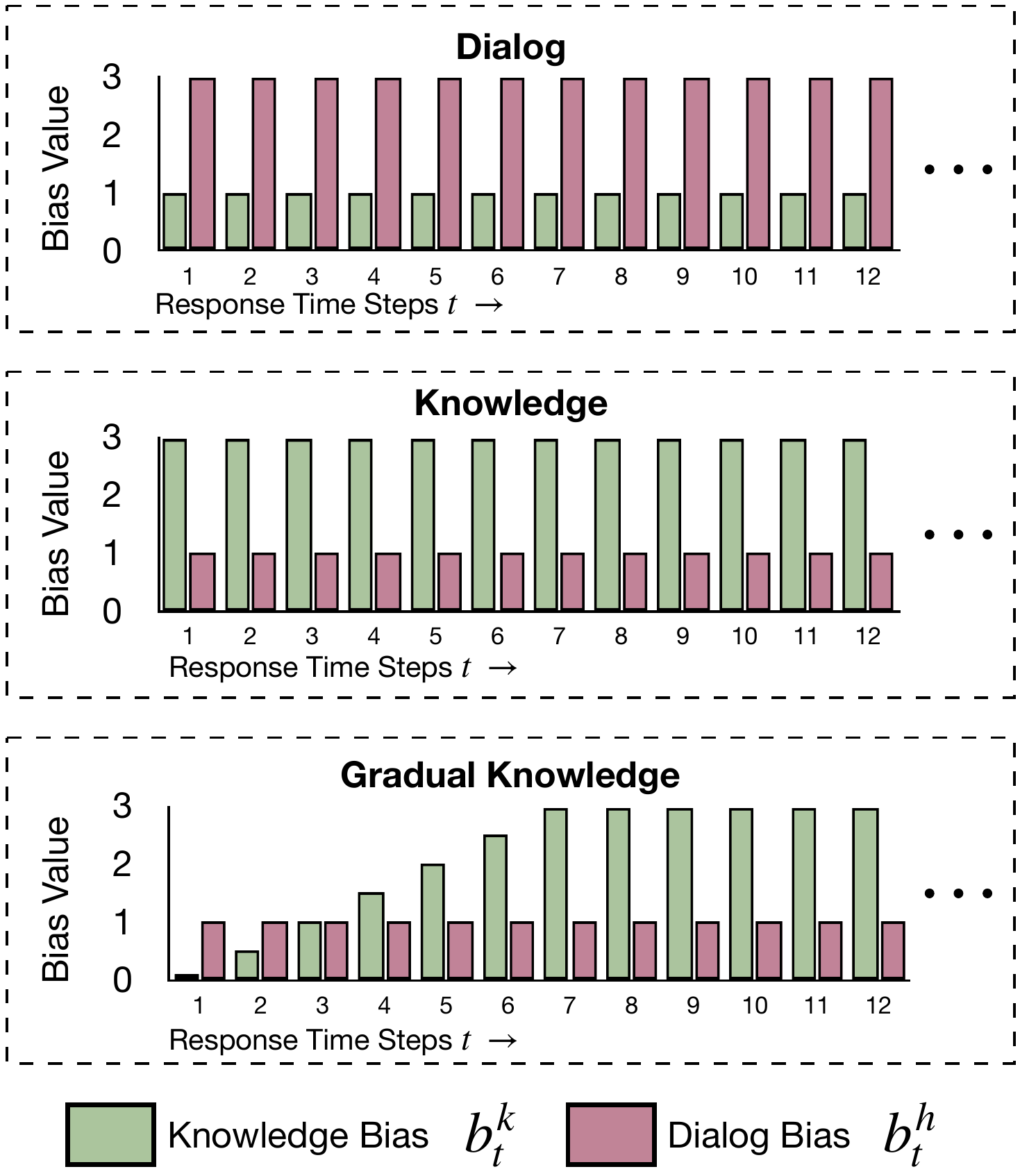}
    \caption{ \footnotesize Cross-attention biasing profiles: Dialog, Knowledge, and Gradual Knowledge (see \Cref{sec:informativeness_bias_profiles}).}
    \label{fig:biasing-profiles}
\end{figure}

\paragraph{Dialog} where the decoder cross-attention is biased towards the dialog history $h$ across all generation time steps. In other words, for all $t$, we set the biases such that $b^h_t$ is larger than $b^k_t$, and more specifically, we set them as $(b^k_t, b^h_t) = (1,5), \; \forall t$.

\paragraph{Knowledge} where the decoder cross-attention is biased towards the knowledge snippet $k$ across all generation time steps. In other words, for all $t$, we set the biases such that $b^k_t$ is larger than $b^h_t$, and more specifically, we set them as $(b^k_t, b^h_t) = (5,1), \;\forall t$.\footnote{All of our biasing profiles are shared across the multiple heads of attention layers. Exploring head-specific biasing is left as a future work.}

\paragraph{Gradual Knowledge}  where initially decoder cross-attention is biased more towards the dialog history, and as the generation time step progresses, the biasing gradually shifts towards the knowledge snippet. The motivation for this design comes from the typical nature of human conversations, where it often is appropriate to start the response by addressing the last utterance of the other party. 
In this biasing profile, knowledge bias value $b^k_t$ increases linearly (with slope $s$) from $0$ up to a certain threshold. Meanwhile the dialog bias is kept at a constant value throughout the generation time steps. For our experiments, we set the parameters of this cross-attention biasing profile as follows: $\max{\left(b^k_t\right)} = 5$, $b^h_t = 1$, and $s = 0.5,$$\;\forall t$. 

\Cref{fig:biasing-profiles} presents sample representations of these three biasing profiles. Note that dialog and knowledge biasing knobs are rather extreme and mimic a gating strategy between knowledge and dialog history. In contrast, the gradual knowledge biasing knob is based on a predominant response structure in human conversations.

\subsubsection{Decoder Mixing Setup}

We explore two profiles for the decoder mixing control knob, both of which use the Topical-Chat fine-tuned EDT-NRG model's decoder and the pre-trained BART's decoder for decoder mixing. For the first profile, we set the decoder mixing vector $\boldsymbol{\alpha}=[0.5, 0.5]$ which performs an averaging operation for the output of pre-trained and fine-tuned decoders at each decoder layer (\Cref{sec:decoder_mixing}). In the second profile, we set $\boldsymbol{\alpha} = [0.7,0.3]$ which gives the $\alpha$ values 0.7 to the Topical-Chat fine-tuned EDT-NRG decoder and 0.3 to the pre-trained BART decoder in order to put more emphasis on generating proper responses and slightly less emphasis on copying from the knowledge snippet. The latter bias profile essentially applies a smaller bias compared to the former one. 

\subsection{Evaluation}
\label{sec:informativeness_evaluation_setup}

To evaluate the generated responses for the dialog context $(k,h)$, we setup both automatic and human evaluations of the responses to measure their \textit{informativeness}, \textit{fluency}, and \textit{relevance}. Due to the high cost of human evaluations we only conduct them for a subset of our experiments.

\paragraph{Informativeness.}
To evaluate informativeness of responses, we use BLEU$_k$, ROUGE$_k$, and METEOR$_k$ as automatic metrics for comparing a generated response with the provided knowledge snippet ($k$) in the dialog context. As automatic metrics on their own are not entirely reliable for evaluation of informativeness of the responses~\cite{DBLP:conf/eacl/BelzR06,DBLP:conf/inlg/LeeGMWK19}, we also perform human evaluation of the generated responses. For this purpose we define a new taxonomy (presented in Table \ref{tbl:informativeness}) over five levels of informativeness. The goal of these levels is to capture the amount of manifestation of provided knowledge in the response. Details on the setup of human evaluation is provided in~\Cref{sec:appendix_informativeness_human_eval}.

\paragraph{Fluency.}
As was mentioned in \Cref{sec:experimental_goals} we also examine if applying the control knobs to create more informative responses would impact the fluency of the generated responses negatively. To that end we set up human evaluations in which annotators make a yes or no decision on the question ``\textit{Does the language of the response seem correct?}''. Moreover, as an automatic metric for fluency, we also measure the perplexity of the models calculated with respect to ground-truth human response (PPL$_r$). In \Cref{sec:appendix_informativeness_additional_metrics}, we discuss additional automatic metrics for fluency.

\paragraph{Relevance.}
To evaluate the relevance of responses, in human evaluation we ask annotators the following question: ``\textit{Regardless of its factual correctness, how appropriate is the response to the conversation?}''. This score is filled on a Likert scale of 1-5.

\begin{table}[t!]
\centering\renewcommand\cellalign{lc}
\setcellgapes{3pt}\makegapedcells
\resizebox{\linewidth}{!}{
    \begin{tabular}{c|l} 
        \textbf{Level} & \textbf{Taxonomy}\\
        \hline
        1 & \makecell{Does NOT include anything from the provided knowledge \\ and does NOT provide any facts.}\\
        2 & \makecell{Does NOT include anything from the provided knowledge \\ but includes some other facts or opinions (made up or not).}\\
        3 & \makecell{Includes some words from the provided knowledge, \\ but makes up facts.}\\
        4 & \makecell{Indirectly uses provided knowledge, without making up facts.}\\ 
        5 & \makecell{Directly uses provided knowledge, without making up facts.} \\
    \end{tabular}
}
\caption{ \footnotesize Proposed taxonomy to evaluate \textit{informativeness} in responses. We prioritize knowledge-oriented responses over un-informative responses (Levels 2-5 vs. 1). Within levels 2-5, we prefer responses that adhere to the provided knowledge (4 and 5) over responses that mention hallucinated facts (2 and 3).}
\label{tbl:informativeness}
\end{table}









\begin{table*}[t!]
    \centering
    \resizebox{\textwidth}{!}{
    \begin{tabular}{|l|l|rr|c|c|ll|ll|ll|c|}
        \hline
        \multirow{3}{*}{\textbf{Knob}} & \multirow{3}{*}{\textbf{Bias Profile}} & \multicolumn{3}{c|}{\textbf{Fluency}} & \textbf{Relevance} & \multicolumn{7}{c|}{\textbf{Informativeness}} \\ \cline{3-13}
        & & \multicolumn{2}{c|}{\textbf{PPL$_{r}$}} & \textbf{Human Eval} & \textbf{Human Eval} & \multicolumn{2}{c|}{\textbf{BLEU$_{k}$}} & \multicolumn{2}{c|}{\textbf{ROUGE$_{k}$}} & \multicolumn{2}{c|}{\textbf{METEOR$_{k}$}} &  \textbf{Human Eval} \\ 
        &&Freq&Rare& $[0,1]$ & $[1,5]$ &Freq&Rare&Freq&Rare&Freq&Rare&$[1,5]$\\
        \hline
        \multicolumn{2}{|c|}{Base Model}  & 9.66 & 9.88 & 0.796 & 3.76  &  0.09 & 0.16 & 0.22 & 0.28 & 0.28 & 0.36 & 3.43 \\
        \hline
        \multirow{3}{*}{A} & \textit{Dialog} & 10.15 & 10.39 & - &  - &  \textbf{0.03} & \textbf{0.10} & \textbf{0.13} & \textbf{0.20} & \textbf{0.16} & \textbf{0.26} & - \\
        & \textit{Knowledge} & 10.20 & 10.59 & 0.786 &  3.81 &  \textbf{0.14} & \textbf{0.26} &	\textbf{0.28} & \textbf{0.38} &	\textbf{0.36} & \textbf{0.49} & \textbf{3.84}\\
        & \textit{Gradual Knowledge}  & 10.03 & 10.38 & 0.788 &  3.83 & \textbf{0.13} & \textbf{0.22} & \textbf{0.26} & \textbf{0.34} & \textbf{0.34} & \textbf{0.45} & \textbf{3.80}\\
        \hline
        B & $\boldsymbol{\alpha} = [0.5, 0.5]$ & 11.78	& 12.02 & 0.785 &  3.87 & \textbf{0.23} & \textbf{0.25} & \textbf{0.35} & \textbf{0.38} & \textbf{0.45} & \textbf{0.48} & \textbf{3.68$^{\dagger}$}\\
        \hline A$+$B&\makecell[l]{\textit{Gradual Knowledge,} \\ $\boldsymbol{\alpha} = [0.5, 0.5]$}  & 12.59 & 13.00 & 0.751 & 3.89 & \textbf{0.28} & \textbf{0.34} & \textbf{0.41} & \textbf{0.47} & \textbf{0.53} & \textbf{0.61} & \textbf{3.82}\\
        \hline
    \end{tabular}
    }
    \caption{ \footnotesize Effect of attention biasing and decoder mixing control knobs on the informativeness of responses on the Topical-Chat \textit{frequent} and \textit{rare} test sets. ``A'' represents the attention biasing knob and ``B'' represents the decoder mixing knob. Results are averaged over 5 inference runs with random seeds. Numbers in boldface (for BLEU$_k$, ROUGE$_k$, METEOR$_k$, and human evaluation) represent statistically significant difference with respect to Base Model as per both pairwise Tukey's HSD test and two-tailed unpaired t-test (both with $p < 0.001$). $^{\dagger}$ Model is significantly different from base for $p < 0.05$ but not for $p < 0.001$. One-way ANOVA test across the five human evaluated models have a statistically significant difference with $p = $1.6e-8 $ < 0.001$. Complete tables, with additional metrics and standard deviations are provided in~\Cref{sec:appendix_informativeness_additional_metrics}, and further discussion on human evaluation setup and results are provided in~\Cref{sec:appendix_informativeness_human_eval}. 
    }
    \label{tab:informative_responses_test}
\end{table*}

\begin{table}[t!]
\centering\renewcommand\cellalign{lc}
\setcellgapes{3pt}\makegapedcells
\resizebox{\linewidth}{!}{
    \begin{tabular}{|l|l|} \hline
        \multirow{11}{*}{\makecell{\textbf{Previous} \\ \textbf{Turns}}} & \textbf{A}: Hello - how are you doing today? \\
        & \makecell[l]{\textbf{B}: Hello, I am doing good. How are you?  \\What do you think of countries having\\ prime ministers?}\\
        & \makecell[l]{\textbf{A}: I'm doing good thanks for asking. \\ I think it's different. What about you?}\\
        & \makecell[l]{\textbf{B}: It is different, I agree.  I am not sure \\ how much power they have.  It seems like they \\ can do a little more than a president can but \\ I am not certain.}\\
        & \makecell[l]{\textbf{A}: I saw the president of the UK has a cat with a \\ government title of chief mouser.}\\
        \hline
        \makecell{\textbf{Knowledge}} & \makecell{\itshape ``Broadly speaking, a ``politician" can be \\ \itshape anyone who seeks to achieve political \\ \itshape power in any bureaucratic institution."} \\ 
        \hline
        \textbf{Response} & \\
        \colorbox{Gray1}{\makecell[r]{Dialog \\ Bias}} & \makecell[l]{\textbf{B}: I wonder how many people are in that \\ position.  I think they can be more than the \\ president though.  They can do whatever \\ they want.}\\
        \colorbox{Blue1}{\makecell[r]{Knowledge \\ Bias \\(Informative)}} & \makecell[l]{\textbf{B}: That is pretty cool. \\ I wonder what kind of job that is. \\ Politicians can be anyone who seeks to achieve \\ political power in any bureaucratic institution. }\\
        \hline
    \end{tabular}
}
\caption{ \footnotesize Example of control between dialog-oriented vs. knowledge-oriented responses. The responses generated with the Dialog and Knowledge biasing profiles, respectively (see ~\Cref{sec:informativeness_bias_profiles}).}
\label{tbl:informative_examples}
\end{table}

\subsection{Results}
\label{sec:informativeness_results}

\Cref{tab:informative_responses_test} summarizes the results of applying attention biasing (Knob A) and decoder mixing knobs (Knob B) for controlling the informativeness of generated responses. 
Note that numbers in boldface represent statistically significant difference from  ``Base Model'', which is the BART-base model fine-tuned on Topical-Chat training set.
In this table, fluency, relevance, and informativeness (column families) of responses generated by applying no control knob (``Base Model'' row), cross-attention biasing knob (rows A), decoder mixing knob (row B), and both attention biasing and decoder mixing knobs (row A+B) are measured. For attention biasing experiments (rows A), we use the three bias profiles that are discussed in \Cref{sec:xatt-profiles}
(Figure \ref{fig:biasing-profiles}).

From the informativeness columns, we can see that using the attention biasing knob for biasing the cross-attention towards dialog (row A - Dialog profile) causes the automatic metrics (BLEU$_k$, ROUGE$_k$, and METEOR$_k$) to drop, indicating that the provided knowledge is incorporated less in the responses, as expected. On the other hand, when the attention biasing knob is used to bias the cross-attention towards the provided knowledge snippet, we see that compared to the base model, these metrics are significantly higher. This trend also appears in the human evaluation, where we see that the informative scores are significantly higher for all the rows corresponding to the attention biasing knob (rows A). Specifically, we see that using the bias profile ``Knowledge'', the human evaluation score for informativeness is 3.84, which is significantly larger than the 3.43 that the base model achieves. 

Regarding fluency and relevance, while we see an increase in perplexity as the model is biased with different profiles, the human evaluations do not show any statistically significant difference between the variants and the base model. This indicates that while the attention biasing knob works well in generating more informative responses, it does not negatively impact the fluency and relevance of the responses. 

Similar conclusions could be made for the decoder mixing knob (row B) as well as the combination of attention biasing and decoder mixing knobs (row A+B). It is notable that for the decoder mixing knob (row B) we see that automatic informativeness scores are higher than those of the attention biasing knobs (rows A), but the human evaluation informativeness score for the attention biasing knobs is higher than that of decoder mixing. This perhaps highlights the value of human evaluation for measuring subjective factors such as informativeness. \Cref{tbl:informative_examples} presents an example from the test set, where we demonstrate how the two cross-attention biasing profiles control the informativeness of the output response.

\begin{table}[t!]
    \centering
    \resizebox{\linewidth}{!}{
    \begin{tabular}{|c|l|cc|cc|}
        \hline
        \multirow{3}{*}{\textbf{Knob}} & \multirow{3}{*}{\textbf{Bias Profile}} &\multicolumn{2}{c|}{\textbf{Knowledge First}} & \multicolumn{2}{c|}{\textbf{Knowledge Last}} \\
        &  & \textbf{PPL}$_r$ & \textbf{ROUGE$_k$} & \textbf{PPL}$_r$ & \textbf{ROUGE$_k$} \\
        \hline
        \multicolumn{2}{|c|}{Base Model}& 9.66 & 0.22 & 9.65 & 0.21\\
        \hline
        B & $\boldsymbol{\alpha} = [0.5, 0.5]$ & 11.78 & 0.35 & 11.38 & 0.17 \\
        \hline 
        A$+$B&\makecell[l]{\textit{Gradual} \\ \textit{Knowledge,} \\ $\boldsymbol{\alpha} = [0.5, 0.5]$} & 12.59 & 0.41 & 11.46 & 0.28 \\
        \hline
    \end{tabular}
    }
    \caption{ \footnotesize Effect of positioning the knowledge snippet in the context: beginning vs. end. We observe that this positioning has an effect on the decoder-mixing knob. Results shown here are on the \textit{frequent} testing set.}
    \label{tab:knowledge_snippet_location}
\end{table}

One point to note here is that since in the experiments knowledge snippets come before the dialog history in the cross-attention context, it is likely that it is easier for the pre-trained decoder in the decoder mixing knob to incorporate the knowledge snippet into the response. If this order is switched and the dialog history comes before the knowledge snippet in the dialog context, using cross-attention biasing in the pre-trained decoder could be leveraged to copy from the knowledge snippet more than the dialog history. We see this effect in~\Cref{tab:knowledge_snippet_location}, where applying only the decoder mixing knob, when knowledge is placed at the end of the dialog context, causes the informativeness to go down, instead of going up. 
The results could be improved using the attention biasing knob along with the decoder mixing knob (row A+B where ROUGE$_k$ increases from 0.17 to 0.28). However, the overall informativeness is significantly lower than the variants where knowledge is placed in the beginning of the dialog context. This indicates that ordering of the input could be crucial for the decoder mixing knob.

\subsubsection{Effect of Bias Amount}

We also examine how the amount of bias in attention biasing (bias value) and decoder mixing knobs ($\boldsymbol{\alpha}$) impact informativeness (through ROUGE$_k$) and fluency (through perplexity) of generated responses. \Cref{tab:biasing_variants} demonstrates the effect of varying the amount of bias of the control knobs. Throughout the experiments on attention biasing (rows A) we keep the value of $b_t^h$ for cross-attention biasing (amount of bias towards dialog history) at 1 in order to measure the effect of different values for $b_t^k$, which represent the amount of bias towards the provided knowledge snippets. From the table, we note that higher values of $b_t^k$ result in higher ROUGE$_k$ which could be interpreted as higher incorporation of knowledge into the generated response. On the other hand, as $b_t^k$ is increased, perplexity of the model also goes up slightly. But if we compare these perplexity numbers with those in Table \ref{tab:informative_responses_test} we can argue that the generated sequences with $b_t^k$ are still reasonably fluent\footnote{Due to high cost of human evaluation, we run them only for a subset of our experiments that cover the primary results.}. 

For rows B in \Cref{tab:biasing_variants} we compare two different decoder mixing profiles, one with $\boldsymbol{\alpha} = [0.7, 0.3]$ and one with $\boldsymbol{\alpha} = [0.5, 0.5]$. The first profile corresponds to the case where we give 0.7 weight to the Topical-Chat fine-tuned decoder and 0.3 to pre-trained BART decoder in decoder mixing at generation time. As is expected, we see that when a higher weight is given to the pre-trained decoder the incorporation of the knowledge snippet into the generated response is higher (higher ROUGE$_k$) which is likely to be due to the higher reconstruction (copying input) behavior of pre-trained BART that results in copying more content from the knowledge snippets into the generated responses. Similar to the attention biasing experiments (rows A), we see that the fluency of the models are likely to be not significantly impacted in these decoder mixing experiments according to the perplexity values.

\begin{table}[t!]
    \centering
    \resizebox{\linewidth}{!}{
    \begin{tabular}{|c|rr|cc|cc|}
        \hline
        \textbf{Knob}& & &\multicolumn{2}{c|}{\textbf{Test frequent}} & \multicolumn{2}{c|}{\textbf{Test rare}} \\
        & $b^k_t$ & $b^h_t$ & \textbf{PPL}$_r$ & \textbf{ROUGE$_k$} & \textbf{PPL}$_r$ & \textbf{ROUGE$_k$} \\
        \hline
        \makecell{None}& 1 & 1 & 9.66 & 0.22 & 9.88 & 0.28 \\
        \hline
        
        \multirow{4}{*}{A}&2 & 1 & 9.78 & 0.25 & 10.05 & 0.32\\
        &5 & 1 & 10.20 & 0.28 & 10.59 & 0.38 \\
        &10 & 1 & 10.70 & 0.32 & 11.18 & 0.41\\
        &50 & 1 & 12.23 & 0.38 & 12.90 & 0.49 \\
        \hline
        \multicolumn{1}{c}{} &  \multicolumn{2}{c}{$\boldsymbol{\alpha}$} & \multicolumn{4}{c}{}\\ 
        \hline
        \multirow{2}{*}{B}& \multicolumn{2}{c|}{[0.7, 0.3]} & 10.47 & 0.24 & 10.68 &  0.31\\
        & \multicolumn{2}{c|}{[0.5, 0.5]} & 11.78 & 0.35 & 12.02 & 0.38\\
        \hline
    \end{tabular}
    }
    \caption{ \footnotesize Effect of varying intensities of biasing for Attention biasing (A) and decoder mixing (B) knobs. For A, we keep the biasing profile to be Knowledge (see ~\Cref{sec:informativeness_bias_profiles}).}
    \label{tab:biasing_variants}
\end{table}

\begin{table}[t!]
    \centering
    \resizebox{\linewidth}{!}{
    \begin{tabular}{|c|c|cc|cc|}
        \hline
        \multirow{3}{*}{\textbf{Knob}} & \multirow{3}{*}{\shortstack[c]{\textbf{Biased} \\ \textbf{Decoder} \\ \textbf{Layers}}} & \multicolumn{2}{c|}{\multirow{2}{*}{\textbf{Test frequent}}} & \multicolumn{2}{c|}{\multirow{2}{*}{\textbf{Test rare}}} \\
         &  & \multicolumn{2}{c|}{} & \multicolumn{2}{c|}{} \\
         & & \textbf{PPL}$_r$ & \textbf{Rouge$_k$} &  \textbf{PPL}$_r$  & \textbf{Rouge$_k$} \\
        \hline
        \multicolumn{2}{|c|}{Base Model} & 9.66  & 0.22 & 9.88 & 0.28 \\
        \hline
        \multirow{3}{*}{A} & Bottom 3  & 9.73  & 0.22 &  9.95 & 0.28 \\
        & Top 3  & 10.11 &  0.28 & 10.48 & 0.37 \\
        & All Layers  & 10.20 & 0.28 & 10.59 & 0.38\\
        \hline
        \multirow{3}{*}{B} & Bottom 3  & 10.23 & 0.20 & 10.45 & 0.25 \\
        & Top 3  &  11.03 & 0.24 & 11.23 & 0.29 \\
        & All Layers  & 11.78 & 0.35 & 12.02 & 0.38 \\
        \hline
        \multirow{3}{*}{A$+$B}& Bottom 3  & 10.77 & 0.29 & 11.05 & 0.36\\
        & Top 3  & 11.58 & 0.34 & 12.01 & 0.42 \\
        & All Layers  & 13.04 & 0.43 & 13.57 & 0.51 \\
        \hline
    \end{tabular}
    }
    \caption{ \footnotesize Effect of varying intensities of biasing through attention biasing (A) or decoder mixing (B) knobs. For A the biasing profile is Knowledge (see ~\Cref{sec:informativeness_bias_profiles}) and for B, $\boldsymbol{\alpha} = [0.5, 0.5]$. }
    \label{tab:layer_variants}
\end{table}

\subsubsection{Layer-Specific Biasing}

Up to this point, we apply the control knobs on all of the transformer decoder layers. Here, we also examine the sensitivity of different layers of the transformer decoder to these knobs. To that end, we consider the settings where the control knobs are applied either to the top half or the bottom half of the transformer decoder layers and compare these to the general setting of applying the knob to all transformer decoder layers. The results are summarized in \Cref{tab:layer_variants}. We can see that for the attention biasing knob (rows A), applying the knob on the bottom three layers has no significant impact on perplexity and ROUGE$_k$ compared to the base model (first row). On the other hand, applying this knob on the top three layers has almost the same effect on perplexity and ROUGE$_k$ as applying the knob to all layers has. The results for the decoder mixing knob (rows B) show that applying this knob only on the bottom three layers deteriorates ROUGE$_k$ compared to the base model. Application of this knob on the top three layers has a very small impact on ROUGE$_k$ compared to the base model, but when this knob is applied to all the layers, we see significant improvement in ROUGE$_k$. Finally, when both of the attention biasing and decoder mixing knobs are applied (rows A+B) to the bottom three, top three, and all the layers, we see that ROUGE$_k$ increases in that same order.

\section{Experiments: Context Augmentation}
\label{sec:control_experiments}

This section explores controlled generation using the context augmentation knob (\Cref{sec:context_biasing_knob}). 
We also investigate how combining the context augmentation knob with attention biasing and decoder mixing knobs could further improve our results.

\subsection{Experiments Setup}
\label{sec:control_experiment_setup}
For demonstrating how the context augmentation knob works and how different parameters and settings impact the results of this knob in ablation studies, in this section, we focus the majority of experiments on controlling the generation of questions (``wh'' questions, yes or no questions, etc.) in dialog responses.  Generating questions is an essential skill towards making dialogs more inquisitive and consequently improving their engagingness with the user~\cite{DBLP:conf/naacl/SeeRKW19}.
In addition to question generation, later in the section, we also show the applicability of the context augmentation knob for controlling other desired attributes in a dialog response, including incorporating feedback-oriented sentences in the response (\Cref{sec:generating_feedback}) and making responses more positive in sentiment (\Cref{sec:sentiment}).


As was discussed in \Cref{sec:context_biasing_knob}, control phrases are used for the context augmentation knob. For applying this knob to generate questions we randomly sample 1000 questions from the Topical-Chat training set and use them as control phrases. We then generate the control code by encoding the control phrases using a pre-trained BART encoder and averaging the encodings over the 1000 samples.


Since in this section we also experiment with combining the context augmentation knob with attention biasing and decoder mixing knobs, we discuss the profiles used for these two knobs here. Regarding attention biasing, unlike the earlier experiments in~\Cref{sec:informativeness_experiments}, the embedding of dialog context $x = (k,h)$ in this section is prepended with the control code $c$ (\Cref{sec:context_biasing_knob}). The resulting augmented dialog context is then segmented as follows. There are two values $b^c_t$ and $b^x_t$ that define the bias vector $\mathbf{b}_t$. In this section we use the profile where $(b^c_t, b^x_t) = (5,1)$ for all $t < 6$ and $(b^c_t, b^x_t) = (1,1)$ for $t \geq 6$, which means that the cross-attention is biased towards the control code for the first $6$ decoder time steps\footnote{The results also hold with other time steps than $6$.}, while there is no cross-attention biasing for the remaining time steps. For the decoder mixing knob, we use $\boldsymbol{\alpha} = [0.5, 0.5]$.

\subsection{Evaluation}
\label{sec:control_experiment_evaluation}

As the initial and ending parts of a dialog typically include greetings and salutations, we sample a subset of test samples from the Topical-Chat test sets by focusing on more central turns in the dialog. In particular, we randomly sample $200$ dialog contexts ($100$ from each \textit{frequent} and \textit{rare} splits of the test set) with five previous dialog turns and use this consolidated test set to evaluate the efficacy of the context augmentation knob. 
For that, similar to~\cite{DBLP:conf/naacl/SeeRKW19}, we use ``?'' as an indicator for questions, which we find to act as a strong proxy for questions\footnote{We rarely find cases where a question does not have a ``?'' or a question-marked sentence is not a question.}.

As each response turn can be composed of multiple sentences, we calculate the number of questions either at the \textit{sentence-level} (counting every sentence with a question mark\footnote{We use NLTK to perform sentence segmentation of the generated response: \protect\url{https://www.nltk.org/_modules/nltk/tokenize.html}.}), or \textit{turn-level} (counting the indicator that a turn has at least one question). We repeat each experiment across five runs, to account for variability in the token sampling procedure. We report both the mean and standard deviation over these counts in the respective tables discussed next. We also measure fluency and relevance through human evaluations similar to the setup that was used earlier in \Cref{sec:informativeness_evaluation_setup}.

\subsection{Results}
\label{sec:control_results}

\Cref{tab:question_control} summarizes the results of biasing the responses towards more questions. C in this table and all the following tables represent the context augmentation knob, and A, B represent cross-attention biasing and decoder mixing knobs, respectively. 
The first row of this table also represents the base case where no biasing knob is applied. From the numbers, one could note that the fluency, that is evaluated by human annotators, does not change much, and any changes are statistically insignificant. For relevance, which is also evaluated by human annotators, although larger differences are observed, they are still not statistically significant\footnote{For Tukey's HSD test between base model `-' and `C+A+B' on \textit{relevance}, we get $p = 0.1994 > 0.05$ which indicates the difference is not statistically significant. These scores have standard deviations of about $1.1$, which could be a reason for no statistically significant difference. Details provided in~\Cref{sec:appendix_control_human_eval}}. The simple intuition that two different responses (one including and the other not including a question) could both be relevant responses, could be the reason why the relevance measure is not significantly impacted, even though significantly more questions are generated in the responses.

In terms of the number of questions generated, we can see that using context augmentation alone (row C) does not generate more questions than the base case (row \textit{None}). However, when this knob is combined with the attention biasing knob (row C+A) or with both attention biasing and decoder mixing knobs (row C+A+B), the number of questions generated is quite larger. From these numbers it appears that cross-attention biasing is key for the context augmentation knob to work. Notably, the combination of context encoding and decoder mixing (row C+B) has the inverse effect of generating fewer questions. This could be due to the pre-trained BART decoder in decoder mixing not being able to use its reconstruction (copying) capabilities on the control codes, which are not sequences of token embeddings because of the averaging operation.

Another notable observation is that despite the control knobs aid in increasing the number of questions by 40\%, not all the responses are becoming questions. This could be indicative of these knobs being used by the model to generate questions whenever it makes sense to have a question in the response, and not forcing the model to generate questions at all times, whether it is appropriate or not; which is a desirable feature of the knobs.


\begin{table}[t!]
    \centering
    \resizebox{\linewidth}{!}{
    \begin{tabular}{|l|c|c|cc|}
        \hline
        \multirow{2}{*}{\textbf{Knobs}} & \textbf{Fluency} & \textbf{Relevance} & \multicolumn{2}{c|}{\textbf{\# of Questions}} \\
        & \makecell{Human \\ Eval} & \makecell{Human \\ Eval} &\textbf{Turn-level}&\makecell{\textbf{Sentence-} \\ \textbf{level}}\\
        \hline
        None & 0.86 & 3.70  & 58.2$\pm$4.2 & 61.4$\pm$5.2 \\
        C & 0.88 & 3.62  & 58.6$\pm$5.4 & 60.0$\pm$6.3 \\
        C$+$A & 0.87 & 3.56 & 70.4$\pm$3.7 & 72.6$\pm$4.1 \\
        C$+$B & 0.85 & 3.61 & 50.2$\pm$3.1 & 59.6$\pm$6.2 \\
        C$+$A$+$B & 0.85 & 3.55 & 83.4$\pm$4.8 & 100.0$\pm$4.7 \\
        \hline
    \end{tabular}
    }
    \caption{ \footnotesize Control over number of generated questions for different combinations of knobs. Configurations of A, B, and C are defined in~\Cref{sec:control_experiment_setup}. Note that sentence-level numbers are higher than turn-level as each turn can have multiple questions. We report the mean and standard deviation across five generations with varying random seeds over the 200 dialog contexts from the consolidated test set (\Cref{sec:control_experiment_evaluation}). Statistical significance details are provided in~\Cref{sec:appendix_control_human_eval}. Qualitative examples are presented in~\Cref{sec:appendix_qualitative_examples}. Reliability of the proxy-based `?' marker for counting questions is established through a human evaluation detailed in~\Cref{sec:appendix_question_detector_human_pref}.}
    \label{tab:question_control}
\end{table}

\subsubsection{Effect of Number of Control Phrases}
\label{sec:effect_of_biasing_set_size}

In the previous experiments, we used a set of 1000 control phrases for the context augmentation knob. In the next experiment, we vary that number  and the results are shown in Table \ref{tab:effect_of_biasing_set_size}. Again, we see that context augmentation alone (row C) and the combination of context augmentation and decoder mixing knobs (row C+B) do not result in more generated questions (at turn level). However, most notably, when the three knobs are combined (row C+A+B) with ten control phrases, the average number of generated questions is 147.4, which is almost three times more than the base case (row \textit{None}). Also, as the number of control phrases increases, the number of generated questions decreases. This could be due to the averaging operator of calculating control codes which might be causing the question aspect of the control phrases not to be the only prominent shared feature between the control phrases.

\begin{table}[t!]
    \centering
    \resizebox{\linewidth}{!}{
    \begin{tabular}{|l|llll|}
        \hline
        \multirow{2}{*}{\textbf{Knobs}}   & \multicolumn{4}{c|}{\textbf{Size of Biasing Set}} \\
        &\textbf{10}&\textbf{100}&\textbf{1K}&\textbf{10K}\\
        \hline
         None &   $56.0 \pm 1.6$ & $55.6 \pm 4.8$ & $58.2 \pm 4.2$ & $53.0 \pm 4.0$ \\
        C &   $63.4 \pm 3.9$ & $59.6 \pm 4.5$  & $58.6 \pm 5.4$ & $59.4 \pm 6.2$ \\
        C$+$A &   $74.2 \pm 2.3$ & $65.2 \pm 4.2$ & $70.4 \pm 3.7$ & $72.2 \pm 4.1$ \\
        C$+$B &   $59.2 \pm 5.8$ & $43.0 \pm 4.1$ & $50.2 \pm 3.1$ & $44.8 \pm 3.6$  \\
        C$+$A$+$B &   $147.4 \pm 2.7$ & $103.4 \pm 5.4$ & $83.4 \pm 4.8$ &  $80.8 \pm 4.6$\\
        \hline
    \end{tabular}
    }
    \caption{ \footnotesize Effect of number of question control phrases on the number of generated questions at turn-level. Configurations of A, B, and C are defined in~\Cref{sec:control_experiment_setup}.}
    \label{tab:effect_of_biasing_set_size}
\end{table}

\subsubsection{Effect of Encoder for Control Phrases}
\label{sec:source_encoder_control}
So far in this section, we have used pre-trained BART's encoder for creating the control codes from control phrases for the context augmentation knob. The next set of experiments show how using Topical-Chat fine-tuned EDT-NRG encoder instead would impact the results. The results of these experiments are shown in Table \ref{tab:source_encoder_control}. We see from these numbers that for cases where the decoder mixing knob is used (rows C+B and C+A+B) using Topical-Chat fine-tuned EDT-NRG encoder for building the control code results in much fewer questions in the generated responses. One potential explanation for this could be that the decoder mixing knob's contributions are hindered by using an encoder that it has not been associated with before. 

On the other hand, when the decoder mixing knob is not involved (rows C and C+A) we see that using the Topical-Chat fine-tuned EDT-NRG encoder is increasing the number of generated questions which again could be due to the familiarity of the decoder with the encoder used for the context augmentation knob.


\begin{table}[t!]
    \centering
    \resizebox{0.7\linewidth}{!}{
    \begin{tabular}{|l|l|l|}
        \hline
        \multirow{2}{*}{\textbf{Knobs}} &  \multicolumn{2}{c|}{\textbf{Biasing Encoder}} \\
        \cline{2-3}
        & \makecell[l]{\textbf{Pre-Trained} \\ \textbf{Encoder}} & \makecell[l]{\textbf{Fine-Tuned} \\\textbf{Encoder}}\\
        \hline
        None &  58.2$\pm$4.2 & 56.4$\pm$6.8 \\
        C &    58.6$\pm$5.4 & 67.0$\pm$2.6 \\
        C$+$A &    70.4$\pm$3.7 & 74.0$\pm$7.2  \\
        C$+$B &   50.2$\pm$3.1  & 41.8$\pm$3.6  \\
        C$+$A$+$B &    83.4$\pm$4.8  & 54.2$\pm$3.8  \\
        \hline
    \end{tabular}
    }
    \caption{ \footnotesize Comparing control over number of generated questions (at turn-level) between pre-trained and fine-tuned encoder for generating control codes. Configurations of A, B, and C are defined in~\Cref{sec:control_experiment_setup,sec:control_experiment_setup}.}
    \label{tab:source_encoder_control}
\end{table}

\subsubsection{Effect of Source of Control Phrases}
\label{sec:source_biasing_phrases}

In this experiment, we evaluate the impact of changing the source of these questions. More specifically, we sample 1000 questions from the SQuAD~\cite{DBLP:conf/acl/RajpurkarJL18} dataset for creating the control code from the context augmentation knob. The results in \Cref{tab:source_biasing_phrases} show that there is no conclusive and significant difference between the two sources (Topical-Chat and SQuAD) of biasing phrase in terms of the final number of generated questions, which suggests that the source of control phrases might not be an important factor, particularly for questions. Moreover, this could also be due to the smoothing out of domain-specific features from the averaging operation in the context augmentation knob.


\begin{table}[t!]
    \centering
    \resizebox{0.9\linewidth}{!}{
    \begin{tabular}{|lc|ll|}
        \hline
        \multirow{2}{*}{\textbf{Knobs}} & \multirow{2}{*}{\shortstack{\textbf{Biasing} \\ \textbf{Encoder}}} & \multicolumn{2}{c|}{\textbf{\# of Questions}}\\
        &&\textbf{Topical-Chat}& \textbf{SQuAD}\\
        \hline
        C &  \multirow{4}{*}{\makecell{Pre-\\Trained\\ BART}}  &  $58.6 \pm 5.4$ & $56.6 \pm 4.4$ \\
        C$+$A &    &  $70.4 \pm 3.7$ & $66.6 \pm 4.0$ \\
        C$+$B &    &  $50.2 \pm 3.1$ & $49.4 \pm 4.2$ \\
        C$+$A$+$B &    &  $83.4 \pm 4.8$ & $114.0 \pm 5.2$ \\
        \hline
        C &   \multirow{4}{*}{\makecell{Fine-\\Tuned\\ BART}} &  $67.0 \pm 2.6$ & $64.4 \pm 5.3$ \\
        C$+$A &   &  $74.0 \pm 7.2$ & $72.8 \pm 4.4$ \\
        C$+$B &   &  $41.8 \pm 3.6$ & $45.0 \pm 5.13$ \\
        C$+$A$+$B &   &  $54.2 \pm 3.8$ & $54.0 \pm 4.14$ \\
        \hline
    \end{tabular}
    }
    \caption{ \footnotesize Comparing control over the number of generated questions (turn-level) between in-domain (Topical-Chat) and out-of-domain (SQuAD) control phrases. Configurations of A, B, and C are defined in~\Cref{sec:control_experiment_setup,sec:control_experiment_setup}.}
    \label{tab:source_biasing_phrases}
\end{table}



\subsection{Context Augmentation for Other Attributes}
In the previous results, we have shown zero-shot controlled generation using the proposed control knobs (specifically the context augmentation knob) for generating questions. One question here is whether such control in generation can be observed for more specific types of questions or over concepts beyond questions, such as other dialog acts like feedback or semantic aspects like sentiment. Next, we explore the answer to these questions.

\subsubsection{Fine-Grained Question Control}
\label{sec:fine-grained-question-control}

In this section, we look into the ability of the control knobs to generate fine-grained question types. We consider the ISO-based Dialog Act Scheme in~\cite{DBLP:conf/coling/MezzaCSTR18}, and in particular, we choose the question types from the subset used in~\cite{DBLP:conf/inlg/HedayatniaGKLEH20}. These include PropQ, ChoiceQ, and SetQ question types. \Cref{tbl:question_types} explains what these three types of questions are using examples. 

\begin{table}[t]
\centering\renewcommand\cellalign{lc}
\setcellgapes{3pt}\makegapedcells
\resizebox{\linewidth}{!}{
    \begin{tabular}{|c|c|c|} \hline
        \makecell{\textbf{Question} \\ \textbf{Type}} & \textbf{Definition} & \textbf{Example} \\
        \hline
        \textit{PropQ} & \makecell[c]{Yes-no\\question} &\makecell{Do you know what the University \\ of Iowa's locker room is?} \\
        \textit{SetQ} & Wh-question &\makecell{What about you?} \\
        \textit{ChoiceQ} & Or-question & \makecell{Or does it become a problem?} \\
        \hline
    \end{tabular}
}
\caption{\footnotesize Fine-grained question types considered for control.}
\label{tbl:question_types}
\end{table}

\paragraph{Evaluation Approach.}

For evaluating the accuracy of generating these fine-grained questions, we initially used the off-the-shelf SVM-based dialog-act classifier proposed in \cite{DBLP:conf/coling/MezzaCSTR18}. However, we found that this model has a slow inference rate, and as a result, we trained an RNN model with a similar training setup as the SVM model. We use this RNN-based model as the primary evaluator of our generated responses. To establish the performance of this model, we conduct human evaluations on a set of $300$ sentences (full details in~\Cref{sec:appendix_control_classifier_human_eval}). The dialog acts tagged by this model achieves F1 score of 0.83, which indicates that this model is a relatively reliable tool for evaluating the generated responses. 

\paragraph{Control Phrases.} 

While it is possible to curate random examples of fine-grained questions from the Topical-Chat training set, we take a different approach here. We sample the most frequent phrases of these question types from the training set and curate small sets of these questions' prefixes. For example, for \textit{PropQ} we curate control phrases that include \textit{``Do you like"}, \textit{``Do you know"}, \textit{``Have you ever"}, \textit{``Are you a"}, etc. The goal of this approach is two-fold. First, we aim to show that we can achieve controlled generation even with a very small set of control phrases. Second, to show that there is no particular requirement for the control phrases to be well-formed questions. As seen in the results below, we observe that incomplete sentences also work as effective control phrases.
 
\paragraph{Results.}

\begin{table}[t!]
    \centering
    \resizebox{\linewidth}{!}{
    \begin{tabular}{|lc|c|c|c|c|}
        \hline
        \multirow{2}{*}{\textbf{Knobs}} & \multirow{2}{*}{\shortstack{\textbf{Biasing} \\ \textbf{Code}}} & \multicolumn{4}{c|}{\textbf{Predictions}}\\
        &&\cellcolor{Gray1}\textbf{PropQ}& \cellcolor{Blue1}\textbf{SetQ}& \cellcolor{almond}\textbf{ChoiceQ} & \cellcolor{ghostwhite}\textbf{Feedback} \\
        \hline
        None & None  & 30.8 & 10.8 & 0.0 & 72.4 \\
        \hline
        C &  \cellcolor{Gray1} & \cellcolor{Gray1}43.4 & 12.0 & 0.0& 64.8 \\
        C$+$A &  \cellcolor{Gray1}  & \cellcolor{Gray1}87.6 & 10.2 & 0.0 & 38.4\\
        C$+$B &   \cellcolor{Gray1} & \cellcolor{Gray1}68.2 &  18.8 & 0.0   & 57.4\\
        C$+$A$+$B & \multirow{-4}{*}{\cellcolor{Gray1}\textit{PropQ}}    &  \cellcolor{Gray1}183.2 & 6.2 & 1.0 & 26.4 \\
        \hline
        C &  \cellcolor{Blue1} & 35.2 & \cellcolor{Blue1}13.2 & 0.0 & 65.0 \\
        C$+$A &  \cellcolor{Blue1}  & 42.8 & \cellcolor{Blue1}23.6 & 0.0 & 56.4\\
        C$+$B &   \cellcolor{Blue1} &  29.0 & \cellcolor{Blue1}34.2 & 0.0 & 59.6 \\
        C$+$A$+$B & \multirow{-4}{*}{\cellcolor{Blue1}\textit{SetQ}} &  37.4 & \cellcolor{Blue1}105.2 & 1.0& 43.4 \\
        \hline
        C &  \cellcolor{almond} &33.6 & 12.4 & \cellcolor{almond}1.0&  67.8\\
        C$+$A &  \cellcolor{almond}  & 50.0 & 15.4 & \cellcolor{almond}0.0& 52.8\\
        C$+$B &   \cellcolor{almond} &  29.8  & 18.8 & \cellcolor{almond}1.5 & 67.0\\
        C$+$A$+$B & \multirow{-4}{*}{\cellcolor{almond}\textit{ChoiceQ}}    &  91.6 & 20.0 & \cellcolor{almond}2.6 & 45.2\\
        \hline
        C &  \cellcolor{ghostwhite} & 31.2 &  9.4 &  0.0 & \cellcolor{ghostwhite}71.0 \\
        C$+$A &  \cellcolor{ghostwhite}  & 33.8 & 10.0 & 0.0 & \cellcolor{ghostwhite}85.6 \\
        C$+$B &   \cellcolor{ghostwhite} &  16.6  & 9.0 & 1.0 & \cellcolor{ghostwhite}97.2 \\
        C$+$A$+$B & \multirow{-4}{*}{\cellcolor{ghostwhite}\textit{Feedback}} &  16.0 & 13.4 & 0.0 & \cellcolor{ghostwhite}141.4 \\
        \hline
    \end{tabular}
    }
    \caption{ \footnotesize Comparing control over the number of questions (at sentence-level) in 200 responses when biased with different fine-grained question types. Configurations of A, B, and C are defined in~\Cref{sec:control_experiment_setup}. Examples of each of the question types and Feedback could be found in \Cref{tbl:control_examples}.}
    \label{tab:fine-grained_questions}
\end{table}

\begin{table}[t!]
\centering\renewcommand\cellalign{lc}
\setcellgapes{3pt}\makegapedcells
\resizebox{\linewidth}{!}{
    \begin{tabular}{|l|l|} \hline
        \multirow{ 5}{*}{\makecell{\textbf{Previous} \\ \textbf{Turns}}} & \textbf{A}: do you watch the NFL?\\
        & \makecell{\textbf{B}: I sure do. How about yourself?}\\
        & \makecell{\textbf{A}: Me too. Did you know women can play \\ in the NFL too?}\\
        & \makecell{\textbf{B}: Did not know there, are there limitations though?}\\
        & \makecell{\textbf{A}: Nope as long as you qualify, \\ you are allowed to play.}\\
        \hline
        \makecell{\textbf{Knowledge}} & \makecell{\itshape ``the NFL uses tracking chips embedded in players \\
        \itshape shoulder pads to record a players every move \\ 
        \itshape on the field. They are also located inside footballs to \\
        \itshape track: velocity, rotation, ball speed and location.''} \\ 
        \hline
        \textbf{Response} & \\
        \colorbox{Gray1}{PropQ} & \makecell{\textbf{B}: \underline{Do you know much about tracking?} Apparently \\ they use tracking chips in the shoulder pads and the  \\ balls to record a player's movements.}\\
        \colorbox{Blue1}{SetQ} & \makecell{\textbf{B}: \underline{What do you think of the tracking chips} \\ \underline{they put in the players?} }\\
        \colorbox{ghostwhite}{Feedback} & \makecell{\textbf{B}: \underline{That is awesome.} Did you know they are putting \\  tracking chips in the players shoulder pads to \\ record their movements?} \\
        \hline
    \end{tabular}
}
\caption{ \footnotesize A sample dialog from the consolidated test set with the controlled generations by Knobs C+A+B. Additional examples (randomly sampled) are provided in~\Cref{sec:appendix_qualitative_examples}.}
\label{tbl:control_examples}
\end{table}

The results are summarized in Table \ref{tab:fine-grained_questions}. We can see that generating PropQ and SetQ questions could be successfully accomplished by using control knobs. Specifically, PropQ questions are generated significantly more compared to SetQ questions. ChoiceQ questions, however, are not being generated using the control knobs. One reason for this could be that such questions are quite rare in the training set of the Topical-Chat, and as a result, the model has not learned how to generate them. The other factor could be that the control phrases for ChoiceQ are not quite representative of what ChoiceQ questions are. A few sample generations for each one of these question types are shown in \Cref{tbl:control_examples}.

In terms of precision of control, from the numbers presented in Table \ref{tab:fine-grained_questions} we can also study how precise the control knobs are. More specifically, we want to determine when the goal is generating more SetQ questions, how much difference is observed in the number of generated PropQ questions. In general, we see that the models largely adhere to the provided additional context. The only place where the precision is poor is ChoiceQ, which means conditioning with ChoiceQ does not improve the number of ChoiceQ questions generated, but it increases the number of generated PropQ questions. This could be due to the similarity between ChoiceQ and PropQ questions in general (see the control phrases in the \Cref{sec:biasing_phrases_example}).


\subsubsection{Generating Feedback Responses}
\label{sec:generating_feedback}

Beyond questions, we show that the proposed control knobs are also effective in creating other dialog acts such as feedback. We evaluate the controllability of feedback acts using the same RNN-based evaluator (\Cref{sec:appendix_control_classifier_human_eval}). In the sampled example in Table \ref{tab:fine-grained_questions}, we can see that using feedback control codes helps with generating significantly more responses that are providing feedback for the previous turn. 

\subsubsection{Sentiment}
\label{sec:sentiment}

We also investigate the use of the context augmentation knob to generate more positive responses. For control phrases we use \textit{``That's awesome''}, \textit{``That's cool''}, \textit{``Oh that is great''}, \textit{``It's great to''}, and \textit{``It's wonderful to''}. The results are shown in \Cref{tab:sentiment_results}. Here again, we see that using the context augmentation knob alone (row C) does not result in statistically significant improvements in the positivity of sentiment of the generated responses (measured using an off-the-shelf sentiment classifier\footnote{\url{https://huggingface.co/transformers/quicktour.html}}). However, similar to the previous experiments, when the context augmentation knob is combined with attention biasing and decoder mixing knobs (row C+A+B), we see the most significant increase in the average sentiment scores. It should be noted that the base model (row \textit{None}) already has a very high average sentiment score (around 0.57) which is indicating that the majority of the responses created by the base model are already positive. This could potentially explain why the increase in the average positivity of the sentiment, although significant, is not very high.

\begin{table}[t!]
    \centering
    \resizebox{0.7\linewidth}{!}{
    \begin{tabular}{|l|c|l|}
        \hline
        \textbf{Knobs} &  $p($positive$|y)$ & p-value \\
        \hline
        None &  0.5697$\pm$0.016 & - \\
        C &   0.5565$\pm$0.007 & 0.1851  \\
        C$+$A &   0.5720$\pm$0.017 & 0.8513 \\
        C$+$B &   \textbf{0.6113$\pm$0.021} & 0.0155  \\
        C$+$A$+$B &  \textbf{0.6508$\pm$0.022} & 0.0005  \\
        \hline
    \end{tabular}
    }
    \caption{ \footnotesize Sentiment scores (1$\rightarrow$positive and 0$\rightarrow$negative) averaged over 5 runs. Models significantly different from base model (based on a two-tailed unpaired t-test) are highlighted using boldface. Configurations of A, B, and C are defined in~\Cref{sec:control_experiment_setup,sec:control_experiment_setup}. Refer to~\Cref{sec:appendix_qualitative_examples} for some qualitative examples.}
    \label{tab:sentiment_results}
\end{table}

\section{Deeper Dive into Attention Mechanisms in Encoder-Decoder Transformers}
\label{sec:analysis}

So far in this work, we have shown the feasibility and efficacy of zero-shot controlled NLG by directly manipulating the internal workings of trained encoder-decoder transformer models at generation time through the proposed control knobs in \Cref{sec:knobs}. Note that this approach to controlled generation does not require any costly training or gradient-based optimization steps during inference. Although we present results for K-NRG, the control knobs could be used for zero-shot control of any EDT-NLG model. 

The counter-intuitive fact that trained encoder-decoder transformer models could go through such drastic manipulations and not only not get fully derailed by them, but also generate sentences with the desired attributes raises many questions. In this section, we try to address some of these questions. Moreover, we believe this observation to have consequences beyond the controlled NLG problem, including more compute efficient approaches towards training these models that we also discuss in this section.

\subsection{Manipulating Self-Attention}
So far we have studied the application of the attention biasing knob on cross-attention modules in encoder-decoder transformer models to control the generation, but we have not yet applied this knob on self-attention modules (D in \Cref{fig:overall}).  

\subsubsection{Self-Attention Biasing}

\begin{table*}[t!]
 \centering
\resizebox{0.9\textwidth}{!}{
    \begin{tabular}{|l|c|l|} \hline
        \multirow{2}{*}{\makecell{\textbf{Self-Attention} \\ \textbf{Mixing}}} &\textbf{Fluency} & \multirow{2}{*}{\textbf{Samples of Model Responses}} \\ 
        &\textbf{(PPL$_r$)}  & \\
        \hline
        \{FT\} & 9.66 & \makecell[l]{ - I agree. I think it's funny that the highest score ever was 222-0. \\ That must have been a humiliating defeat.}\\
        \hline
        \hline
        \{PT\}  & 10.52 & \makecell[l]{- I do like the Patriots. What about you?} \\ 
        \hline
        \hline
        \{FT,PT\} & 9.84 & \makecell[l]{- I did not know that. I wonder if they are allowed \\ to eat in restaurants.}\\
        \hline 
        \{FT$_1$,FT$_2$,FT$_3$,FT$_4$\} & 9.68 &  \makecell[l]{- Not really, I think they have some pretty good movies, I don't know.} \\ 
        \hline
        \{FT$_1$,FT$_2$,FT$_3$,FT$_4$,PT\}  & 9.65 &  \makecell[l]{- I did not know that! I really love the batman character. Did you know \\ he was originally named Bat-Man?}\\ 
        \hline
    \end{tabular}
}
\caption{ \footnotesize Mixing multiple self-attention decoder blocks.  Here, FT represents a fine-tuned self-attention block and PT represents the pre-trained self-attention block. For each decoder layer, we perform a convex combination of the participating self-attention functions as per~\Cref{fig:self-att-mix}. The last column represents generated responses (randomly selected).}
\label{tbl:self-attn-mixing}
\end{table*}

We investigated the question of whether the attention biasing knob could be applied to self-attention in a similar way that it was successfully applied to cross-attention. Through our experiments we found that the answer to this question is probably negative. For instance, in a series of experiments we tried to use the attention biasing knob on self-attention modules of the decoder so that the model pays more attention to tokens immediately preceding the present generation time step\footnote{We applied a linear decay bias that from $1$ to $0$ for preceding time steps $t-1$, $t-2$, \dots. This profile in some aspects is similar an n-gram language model where token $y_{t+1}$ is primarily conditioned on its immediately preceding n-gram tokens.}, and we notice that this would cause the generated sentences to be not fluent anymore. More concretely, the average perplexity of generated sentences, measured using a pre-trained GPT-2 model used as a proxy for fluency, is 142.6 compared to the same model with no self-attention biasing that gives the average perplexity of 40.2. This increase in perplexity shows itself as many grammatical and syntactical mistakes in the generated sentences. 

This experiment along with several other similar failed experiments that we ran on biasing the self-attention modules in the decoder of encoder-decoder transformer models raises the hypothesis that perhaps decoder self-attention in these models is primarily responsible for fluency of the generated sentences, and that is why manipulation of self-attention results in loss of fluency. More formally: 
\begin{hypothesis}
\label{hyp:1}
In EDT-NLG models, fluency of the generation is managed by decoder self-attention.
\end{hypothesis}

It should be noted that the observation in \Cref{sec:informativeness_experiments} that biasing cross-attention, while decoder self-attention remains intact, does not negatively impact fluency of generations is another strong evidence for this hypothesis to be correct.

\subsubsection{Self-Attention Mixing}

Inspired by \Cref{hyp:1}, one could ask whether decoder self-attention in EDT-NLG works independently of the task that the model is trained on. In other words, would it be possible to replace the self-attention modules of one trained EDT-NLG model with those of another trained EDT-NLG model and still get fluent generation out of these models?

To examine this, we take two BART models, one is fine-tuned for the Topical-Chat task (here is referred to as fine-tuned BART), and the other is the original pre-trained BART model. Note that the fine-tuned BART is trained to generate responses for a given dialog history and a knowledge snippet, whereas the pre-trained BART is trained to reconstruct an input sentence. 
We replace the parameters of the decoder self-attention of fine-tuned BART with the parameters of the pre-trained BART, and we generate for the Topical-Chat task using the resulting model. The result is generated responses that are surprisingly fluent (row \{PT\} in \Cref{tbl:self-attn-mixing}) with perplexity 10.53 which is only slightly higher than the perplexity of the Topical-Chat fine-tuned model which is 9.66 (row \{FT\} in \Cref{tbl:self-attn-mixing}). It is important to note that these self-attention modules are significantly different from one another, in that the average Frobenius norm of the difference between Q, K, and V matrices are 5.25, 5.60,  and 5.19, respectively, where the average norm of the matrices are 61.10, 61.15, and 33.56, respectively. This could be interpreted as significant difference between the two self-attention modules. This result is quite surprising in that we are replacing all of the parameters of self-attention modules of one trained encoder-decoder transformer model (EDT-NRG for Topical-Chat) with the parameters of another model that has an identical architecture but is trained for a completely different task, and we still see that the performance of the resulting model is not impacted significantly.

\begin{figure}[t!]
    \centering
    \includegraphics[width=0.48\textwidth]{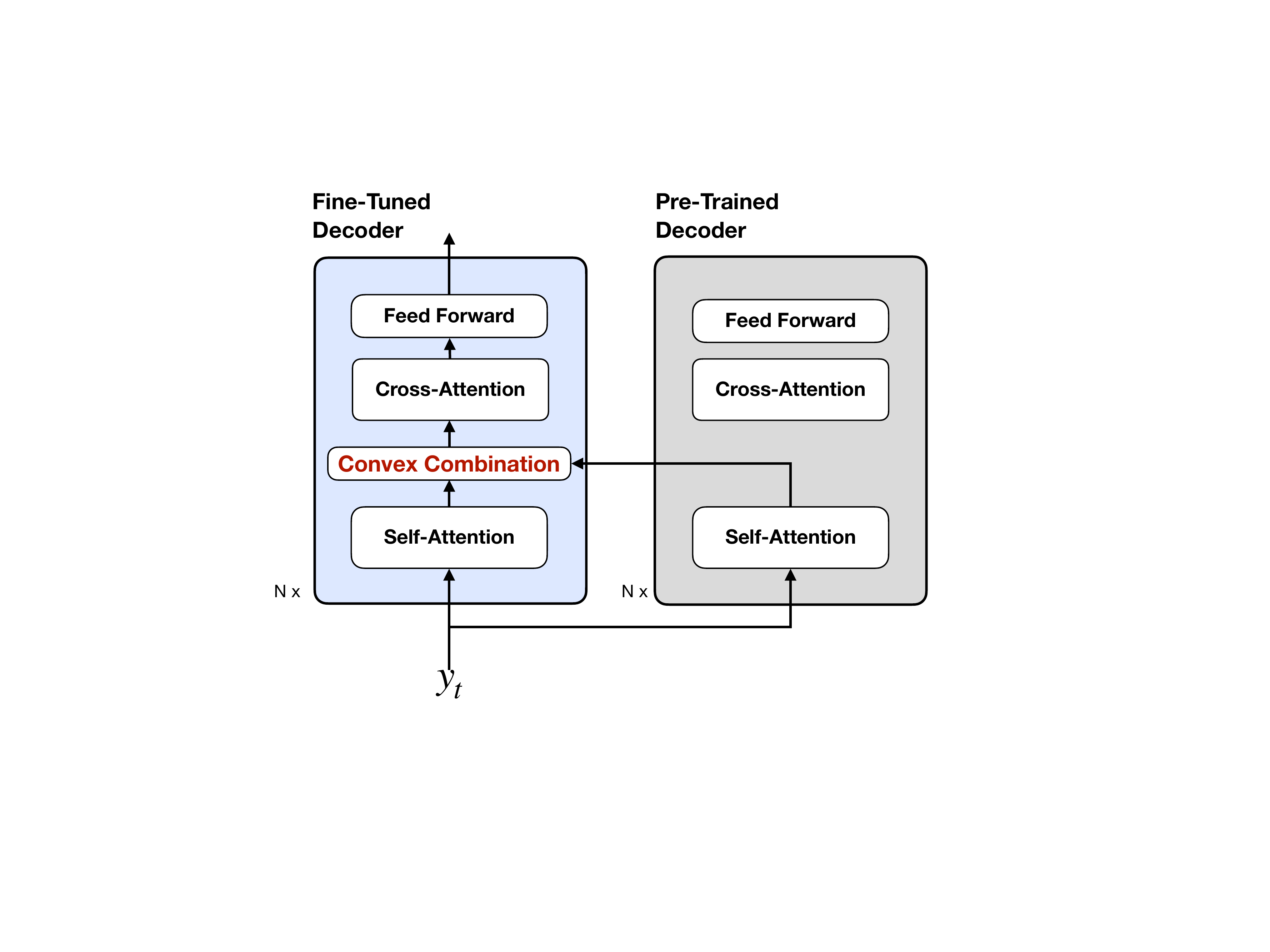}
    \caption{\footnotesize Mixing self-attention of fine-tuned BART for Topical-Chat dataset and pre-trained BART }
    \label{fig:self-att-mix}
\end{figure}

Next we examine the perplexity of generations for the case where the self-attention modules for fine-tuned BART and pre-trained BART are combined. The architecture is shown if \Cref{fig:self-att-mix}. In this architecture at every generation time step two different self-attention modules (one from Topical-Chat fine-tuned BART and one from pre-trained BART) are run and the results are combined through a convex combination. The combined self-attention output then goes through the rest of layers of fine-tuned BART. From the results, shown in row \{FT,PT\} in \Cref{tbl:self-attn-mixing}, we see that the average generation perplexity in this setting is 9.84 which is very close to this metric for fine-tuned BART, which indicates that the model is generating fluent responses.

We next combine four different (trained with different random seeds) fine-tuned BART models' self-attention modules (row \{FT$_1$,FT$_2$,FT$_3$,FT$_4$\} in \Cref{tbl:self-attn-mixing}), and also four different fine-tuned BART and pre-trained BART models' self-attentions modules (row \{FT$_1$,FT$_2$,FT$_3$,FT$_4$, PT\} in \Cref{tbl:self-attn-mixing}), in the same way that is depicted in \Cref{fig:self-att-mix}. We see that the generations from both of these models have a perplexity that is very close to fine-tuned BART's generations perplexity, which again is an indicator that these models with combined self-attention modules generate fluently. It should be noted that the self-attention modules of the four different fine-tuned BART models are significantly different from each other in that the average Frobenius Norm of the difference between Q, K, and V matrices are 6.79, 6.98, and 6.29 respectively\footnote{The average is calculating among differences of pairs of the same matrices from different models}.





We can see from the results that convex combinations (average to be more specific) of decoder self-attention from models that are trained to generate fluent sentences also generates fluent sentences. Moreover, adding random noise to a decoder self-attention module that is trained to generate fluent sentences results in large hits to the fluency of the generated results. These observations would establish additional evidence for Hypothesis \ref{hyp:1}. Also, fluency of convex combination of trained decoder self-attention suggests that, intuitively speaking, there might be a flat surface of fluency in the space of parameters of these encoder-decoder transformer models. It is important to note that if the encoder-decoder transformer architecture is replaced with a transformer decoder architecture, majority of what was discussed above will not hold true. As to why, it is important to notice that for a transformer decoder trained on the Topical-Chat dataset, the decoder self-attention is responsible both for maintaining  fluency of the generated response as well as its relevance to the previous turns of the conversation.

\subsection{Rethinking Transformer Decoder Architectures}
If Hypothesis \ref{hyp:1} holds, one question that arises is are there alternative layouts for transformer decoder that could better facilitate this separation of roles between decoder self-attention and decoder cross-attention? Another natural question here is that if the hypothesis is correct and in a pre-trained encoder-decoder transformer architecture decoder self-attention is already able to generate fluently, can we freeze self-attention parameters during fine-tuning of these pre-trained models? In this section we address these two questions.

\subsubsection{Parallel Self- and Cross-Attention for Transformer Decoders}

\begin{figure}[t!]
    \centering
    \includegraphics[width=0.9\linewidth]{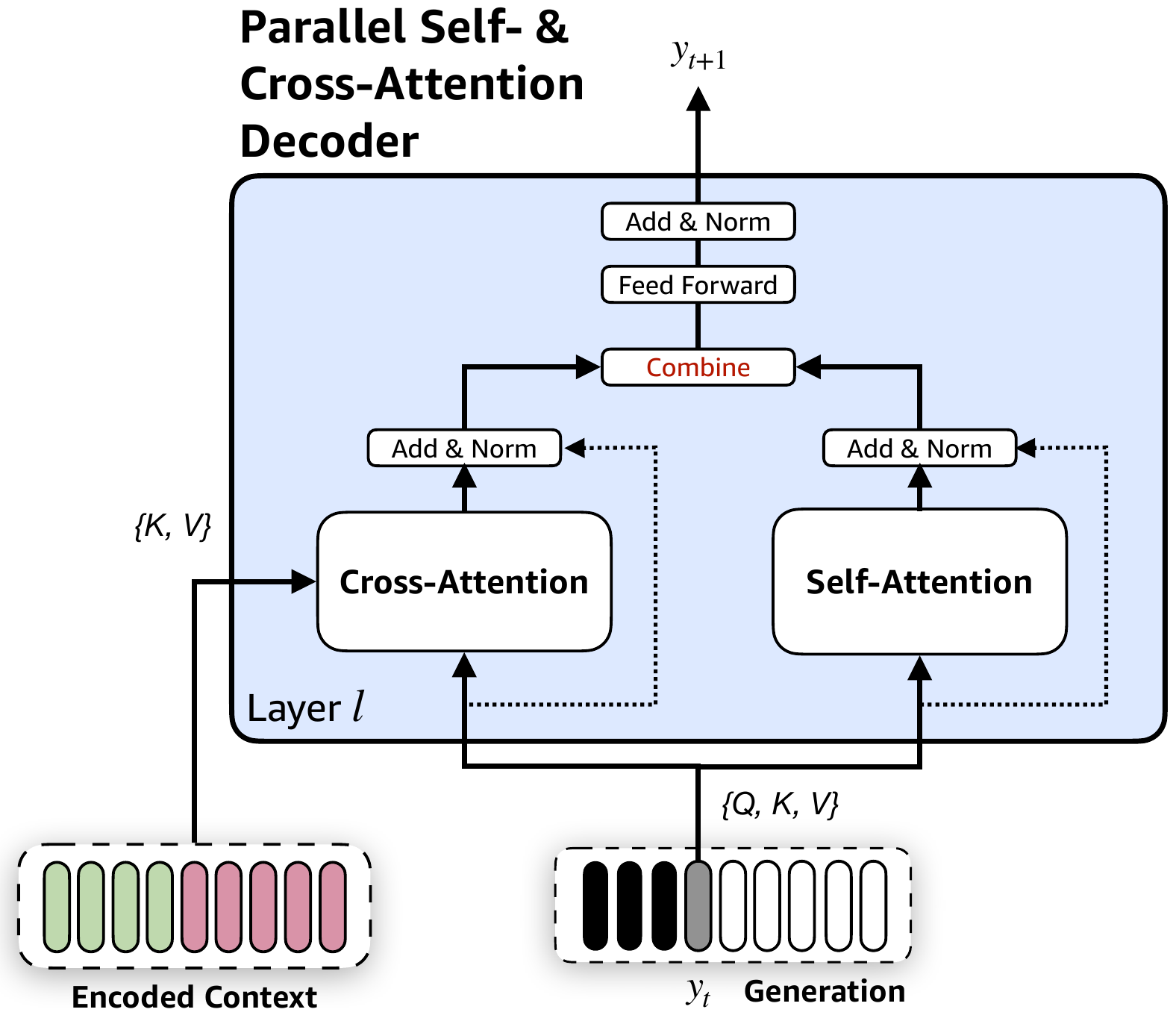}
    \caption{ \footnotesize Parallel self- and cross-attention in a transformer decoder layer.}
    \label{fig:parallel-att}
\end{figure}

\begin{table}[t!]
\centering
\resizebox{\linewidth}{!}{
    \begin{tabular}{|l|cc|c|} 
        \hline
        \textbf{Decoder} & \multicolumn{2}{c|}{\textbf{PPL$_r$}} & \textbf{Sample Responses} \\
        & Freq & Rare & \\
        \hline 
        Sequential & 9.66 & 9.88 & -\\
        \hline
        Parallel & 10.22 & 10.76 & \makecell[l]{- It was released in 2017. I was excited \\ to see it. I am excited to see what the \\ new Avengers movie will be about. \\ 
        - He has been good for a long time. \\
        I think he is going to join the Giants.}\\
        \hline
    \end{tabular}
    }
    \caption{ \footnotesize Parallel decoder self- and cross-attention EDT-NRG performance on the Topical-Chat Dataset.}
    \label{tbl:parallel-decoder-results}
\end{table}

In the current architecture of transformer decoder self-attention, cross-attention, and feed forward layers are sequentially chained together across multiple layers of stacked transformer decoders (\Cref{fig:overall}). If Hypothesis \ref{hyp:1} holds and, as a result, self-attention and cross-attention models in encoder-decoder transformer models have different roles, could a different architecture, in which decoder self-attention and cross-attention are linked not sequentially but in parallel, work better for NLG tasks? The intuition behind this alternative architecture would be somewhat adding more separation between self-attention and cross-attention in the sense that the output of one is not directly the input of the other. To examine this idea, we take a pre-trained BART and change the architecture of its transformer decoder so that the self-attention and cross-attention modules are linked together in a parallel manner instead of the original sequential way. Figure \ref{fig:parallel-att} shows this alternative transformer decoder architecture. Note that we use the original pre-trained BART values, which were trained in the standard sequential decoder, for the parameters of both self- and cross- attention modules for initialization. We then fine-tune this model with the Topical-Chat task. \Cref{tbl:parallel-decoder-results} presents the results where the model does not observe a significant degradation in terms of perplexity. These results along with some qualitative review of the generations suggest positive indications towards the feasibility of such alternate parallel architectures for transformer decoders.

\subsubsection{Fine-Tuning Only Cross-Attention}

Trained EDT-NLG models could generate fluently. If Hypothesis \ref{hyp:1} holds and fluency of the output in these models is managed by decoder self-attention, then if we freeze the decoder self-attention parameters during fine-tuning of these models for other NLG tasks (e.g., NRG for Topical-Chat), would the fine-tuned model still perform well? In a set of experiments, we study this question. More specifically, we freeze all parameters of the transformer decoder except the cross-attention parameters of a pre-trained BART model and we fine-tune the rest of the parameters on the Topical-Chat dataset. The top half of Table \ref{tab:efficient_training} presents the results and compares it with the case where all of the pre-trained BART model are fine-tuned (first row). In the case that decoder self-attention parameters of pre-trained BART (all decoder self-attention parameters except for cross-attention parameters) are fixed and the rest of the parameters are fine-tuned (second row) we see that the impact on perplexity is very small\footnote{We train three independent models with random seeds for both variants. According to two-tailed unpaired t-test, the difference is not statistically signification (p-values are 0.11 and 0.25 for frequent and rare test sets, respectively)}. 

For the third and fourth rows of the table, the parameters of the decoder self-attention of pre-trained BART are replaced with random values. We see that random self-attention parameters when fixed during training cause very high perplexity values (fourth row) and the generations are no longer fluent. This experiment was conducted to ensure that if Hypothesis \ref{hyp:1} is correct, not any random decoder self-attention could result in fluent generations. On the other hand, when the decoder self-attention parameters of pre-trained BART are set to random values, but they are trained during fine-tuning (third row), fluency of the model to some extend is regained.

\begin{table*}[t]
    \centering
    \resizebox{0.9\linewidth}{!}{
    \begin{tabular}{|c|clcc|ll|}
        \hline
        \textbf{Model} &  \makecell{\textbf{Total} \\ \textbf{Parameters}}& \makecell{\textbf{Trainable} \\ \textbf{Parameters}} & \makecell{\textbf{Randomly} \\ \textbf{Initialized}} & \makecell{\textbf{Decoder }\\ \textbf{Fine-tuning}} & \multicolumn{2}{c|}{\textbf{PPL} ($\downarrow$ better)} \\
        & & & \textbf{Dec-Self-Attn.} &  & (\textbf{Test-freq}) & (\textbf{Test-rare}) \\
        \hline
        \hline
        \multirow{2}{*}{Bart-large} & \multirow{2}{*}{406M} & 406M & - & full & 9.31$\pm$0.05 & 9.37$\pm$0.02 \\
        & & 254M (\textcolor{darkgreen}{$\downarrow$37.43\%}) &- & only cross-attn. & 9.40$\pm$0.06 & 9.40$\pm$0.01 \\
        \hline
        \hline
        \multirow{2}{*}{Bart-large} & \multirow{2}{*}{406M} & 406M & $\checkmark$ & full & 11.29 & 11.51 \\
        & & 254M (\textcolor{darkgreen}{$\downarrow$37.43\%}) & $\checkmark$ & only cross-attn. &  18.07 & 17.93 \\
        \hline
        \hline
        Bart-large & \multirow{2}{*}{381M}  & 381M (\textcolor{darkgreen}{$\downarrow$6.15\%}) & - & full & 9.71 & 9.88 \\
        (top-6) & & 228M (\textcolor{darkgreen}{$\downarrow$43.84\%}) & - & only cross-attn. & 12.7 & 12.9 \\
        \hline
        Bart-large & \multirow{2}{*}{381M} & 381M (\textcolor{darkgreen}{$\downarrow$6.15\%}) & - & full & 10.78 & 11.65 \\
        (bottom-6) & & 228M (\textcolor{darkgreen}{$\downarrow$43.84\%}) & - & only cross-attn. & 25.08 & 26.97 \\
        \hline
        Bart-large & \multirow{2}{*}{381M}  & 381M (\textcolor{darkgreen}{$\downarrow$6.15\%}) & - & full & 9.88 & 10.14 \\
        (alternate-6) & & 228M (\textcolor{darkgreen}{$\downarrow$43.84\%}) & - & only cross-attn. & 10.32 & 10.40 \\
        \hline
    \end{tabular}
    }
    \caption{ \footnotesize Training BART-large models with varying initialization (of decoder's self-attention) and decoder self-attention freezing (freezing all decoder parameters except cross-attention and shared embedding matrices) strategies. All $\downarrow/\uparrow$ are relative changes with respect to the BART large model size.}
    \label{tab:efficient_training}
\end{table*}

\subsubsection{Removing Some of the Decoder Cross-Attention Modules}
\label{sec:efficient_training}

We also apply another set of modifications to the BART model in which we remove the cross-attention of some of the transformer decoder layers. In 3 different settings we remove the cross-attention for top half of the decoders, bottom half of the decoders, and every other decoder. The results of these settings are presented in the bottom half of Table \ref{tab:efficient_training}. For the case where only the top half of the decoder layers keep their cross-attention modules (``top-6'' rows in Table \ref{tab:efficient_training}) we see that when all the parameters are fine-tuned (first row of ``top-6'' rows), the model still  performs quite well when compared to the original model. Remember that these models have approximately 8\% less trainable parameters compared to the original BART model. In this setting when all but cross-attention weights on the decoder side are not fine-tuned (second row of ``top-6'' rows), perplexity goes even higher, but is still somewhat acceptable. When only the bottom half of the transformer decoders keep their cross-attention (``bottom-6'' rows in Table \ref{tab:efficient_training}), the performance of the fine-tuned model (first row of ``bottom-6'' rows) model is worse than the case where top half of the decoder transformers kept their cross-attention, but the performance remains in the acceptable range. However, in the case where decoder self-attention parameters are also fixed during training (second row of ``bottom-6'' rows), we see a large hit to the performance and the generations are no longer fluent. Finally in the case where every other transformer decoder keeps its cross-attention (``alternate-6'' rows in Table \ref{tab:efficient_training}) produces the most interesting results. We can see that in this case even when decoder self-attention parameters are fixed during fine-tuning the perplexity on the test-rare set is 10.40 which is only slightly higher than 9.36 which is for the case where all the transformer decoder cross-attention modules are kept and all the parameters are fine-tuned. This result is interesting because in this case on the decoder side none of the self-attention parameters are trained; nor are half of the cross-attention parameters are even in the model. In fact, in this case the number of trainable variables is only 56\% of the trainable variables in the base model and the model's performance is strikingly high.

\subsubsection{Efficient Training of Encoder-Decoder Transformer Models}

The numbers in  Table \ref{tab:efficient_training} suggest that there could be more efficient ways of training EDT models for NLG applications. In the experiments that are summarized in this Table, we see that on the one hand, freezing parameters of decoder self-attention modules to pre-trained values does not hugely impact the performance of the model. Note that this finding is aligned with Hypothesis \ref{hyp:1}. Freezing these parameters during training (or fine-tuning) would mean that gradients for these parameters need not to be kept, tracked, or communicated between GPUs or comput nodes, which would result in significant savings in compute resources. On the other hand, dropping cross-attention modules from some of the transformer decoders would also result in the same savings and reduction in the model size, which results in savings during both training and inference time.


\section{Conclusion}
\label{sec:conclusion}

In this work, we propose novel approaches to controlling NLG models that are based on encoder-decoder transformers. In these approaches, we manually intervene in the internal computations of these EDT-NLG models at generation time to achieve the control goals in a zero-shot manner. These manual interventions are applied through three proposed control knobs: attention biasing, decoder mixing, and context augmentation knobs. Some aspects of applying these knobs on the EDT-NLG models are quite counter-intuitive. Most prominently, the fact that we can manually intervene in computations of these NLG models in rather intrusive ways without derailing the generation process entirely, comes as a surprise. Building on this observation, we then see that in most cases, intuition-based design of manual interventions produce results that are aligned with the intuition behind the design. 

One notable aspect of the results of experiments on the application of these knobs is that using the combination of these knobs leads to the most favorable results. This was specifically pronounced for the context augmentation knob, where applying it alone would result in little to no control in the generation process. However, when combined with the other two knobs, it would result in a large increase in controllability. The context augmentation knob could be thought of as an alternative way of prompting generative language models~\cite{brown2020language}. But it is known that prompting fails to achieve the control goals in models that are not enormous in size~\cite{DBLP:journals/corr/abs-2009-07118}. For instance, we know that prompting works very well for GPT-3, but not necessarily for GPT-2. In this work, we show that by adding cross-attention biasing and decoder mixing to the context augmentation knob (which could be thought of as an alternative to prompting) we can achieve zero-shot controllability for models that are orders of magnitude smaller than GPT-3 (e.g., BART-base).

When used for controlling the generation, we see that while applying the attention biasing knob on cross-attention achieves the desired outcome, it turns out that applying this knob on self-attention results in loss of fluency. That leads us to the hypothesis that decoder self-attention in EDT-NLG models is responsible for fluency of the generations. We examine this hypothesis in several ways, and all the evidence points towards the hypothesis being correct. Inspired by this hypothesis, we propose alternative architectures for transformer decoders that are significantly more compute efficient during both training and generation.

In this work, we show given a control goal, how to generate according to the goal in a zero-shot fashion. One obvious direction for future research is how to develop these control goals, especially in the context of dialog systems, and building models that can both determine the control goals and generate according to them in an end-to-end fashion. As another future research direction, more computational studies are needed to determine how to benefit from the proposed results in designing more compute-efficient encoder-decoder transformer models. 

Application of transformers, and attention in general, goes far beyond NLG and even NLP, and these mechanisms are heavily employed in machine vision, multi-modal learning, and more. Zero-shot biasing of attention through the attention biasing knob that is introduced in this work could potentially be useful in these areas not only for generation, but also other tasks such as classification, segmentation, etc. Also, it should be noted that the attention biasing knob is not limited to the attention mechanism in the context of transformers and could be applied to any attention mechanism, whether it is part of a transformer or not.




\section{Acknowledgements}
We are truly grateful for valuable feedbacks from Gokhan Tur, Yang Liu, Behnam Hedayatnia, Nicole Chartier, and Mohit Bansal.

\bibliographystyle{acl_natbib}
\bibliography{references}

\appendix

\begin{table*}[ht!]
    \centering
    \resizebox{\textwidth}{!}{
    \begin{tabular}{|l|l|llllll|llll|}
        \hline
        \multirow{2}{*}{\textbf{Knob}} & \multirow{2}{*}{\textbf{Bias Profile}} & \multicolumn{6}{c|}{\textbf{Human Response}} & \multicolumn{4}{c|}{\textbf{Knowledge}} \\
        & & \textbf{PPL$_{r}$} & \textbf{F1$_{r}$} & \textbf{BLEU$_{r}$} & \textbf{ROUGE$_{r}$} & \textbf{METEOR$_{r}$} & \textbf{B-Score$_{r}$} & \textbf{F1$_{k}$} & \textbf{BLEU$_{k}$} & \textbf{ROUGE$_{k}$} & \textbf{METEOR$_{k}$} \\
        \hline
        None &  & 9.66 & 0.30 & 0.04 & 0.21 & 0.23 & 0.27 & 0.31 & 0.09 & 0.22 & 0.28 \\
        \hline
        \multirow{3}{*}{\textbf{A}} & \textit{Dialog} & 10.15 & 0.26 & 0.03 & 0.18 & 0.20 & 0.24 & 0.196 & 0.033 & 0.134 & 0.161 \\
        & \textit{Knowledge} & 10.20 & 0.30 & 0.04	& 0.21 & 0.24 & 0.27 & 0.39 & 0.14$\pm$0.01 &	0.28$\pm$0.01 &	0.36$\pm$0.01 \\
        & \makecell[l]{\textit{Gradual}\\\textit{Knowledge}}  & 10.03 & 0.30 & 0.04 & 0.21 & 0.24 & 0.27 & 0.37 & 0.13 & 0.26 & 0.34 \\
        \hline
        \textbf{B} & \makecell[l]{$\boldsymbol{\alpha} = [0.5, 0.5]$} & 11.78	& 0.27 & 0.03 & 0.18 & 0.24 & 0.23 & 0.43 & 0.23$\pm$0.04 &	0.35$\pm$0.04 & 0.45$\pm$0.05 \\
        \hline
        \textbf{A$+$B}&\makecell[l]{\textit{Gradual}\\\textit{Knowledge}\\ $+$ $\boldsymbol{\alpha} = [0.5, 0.5]$}  & 12.59 & 0.28 & 0.03 & 0.19 & 0.25 & 0.23 & 0.49 & 0.28 & 0.41 & 0.53 \\
        \hline
        \multicolumn{11}{l}{$^{\star}$B-Score $\rightarrow$ Bert-Score;}
    \end{tabular}
    }
    \caption{ \footnotesize Effect of Control Knobs on the informativeness of responses on the \textit{Topical-Chat test-freq}. Results are averaged over 5 inference runs with random seeds. For brevity, we report the standard deviation only when it is $>0.01$.}
    \label{tab:appendix_informative_responses_test_freq}
\end{table*}

\begin{table*}[ht!]
    \centering
    \resizebox{\textwidth}{!}{
    \begin{tabular}{|l|l|llllll|llll|}
        \hline
        \multirow{2}{*}{\textbf{Knob}} & \multirow{2}{*}{\textbf{Bias Profile}} & \multicolumn{6}{c|}{\textbf{Human Response}} & \multicolumn{4}{c|}{\textbf{Knowledge}} \\
        & & \textbf{PPL$_{r}$} & \textbf{F1$_{r}$} & \textbf{BLEU$_{r}$} & \textbf{ROUGE$_{r}$} & \textbf{METEOR$_{r}$} & \textbf{B-Score$_{r}$} & \textbf{F1$_{k}$} & \textbf{BLEU$_{k}$} & \textbf{ROUGE$_{k}$} & \textbf{METEOR$_{k}$} \\
        \hline
        None &  & 9.88 & 0.31 & 0.05 & 0.21 & 0.25 & 0.27 & 0.38 & 0.16 & 0.28 & 0.36 \\
        \hline
        \multirow{3}{*}{\textbf{A}} & \textit{Dialog} & 10.39 & 0.27$\pm$.02 & 0.04$\pm$.01 & 0.19 & 0.22$\pm$.02 & 0.24 & 0.28$\pm$.11 & 0.10$\pm$.08 & 0.20$\pm$.08 & 0.26$\pm$.11 \\
        & \textit{Knowledge} & 10.59 & 0.31 & 0.06 & 0.22 & 0.26 & 0.27 & 0.50 & 0.26 & 0.38 & 0.49 \\
        & \makecell[l]{\textit{Gradual}\\\textit{Knowledge}}  & 10.38 & 0.32 & 0.06 & 0.22 & 0.26 & 0.27 & 0.46 & 0.22 & 0.34 & 0.45 \\
        \hline
        \textbf{B} & \makecell[l]{$\boldsymbol{\alpha} = [0.5, 0.5]$} & 12.02 & 0.29 & 0.05 & 0.19 & 0.25 & 0.24 & 0.44 & 0.25 & 0.38 & 0.48 \\
        \hline
        \textbf{A$+$B}&\makecell[l]{\textit{Gradual}\\\textit{Knowledge}\\ $+$ $\boldsymbol{\alpha} = [0.5, 0.5]$}   & 13.00 & 0.29 & 0.05 & 0.20 & 0.27 & 0.23 & 0.54 & 0.34 & 0.47 & 0.61\\
        \hline
        \multicolumn{11}{l}{$^{\star}$B-Score $\rightarrow$ Bert-Score;}
    \end{tabular}
    }
    \caption{ \footnotesize Effect of Control Knobs on the informativeness of responses on the \textit{Topical-Cht test-rare}. Results are averaged over 5 inference runs with random seeds. For brevity, we report the standard deviation only when it is $>0.01$.}
    \label{tab:appendix_informative_responses_test_rare}
\end{table*}

\section{Model Details}
\label{sec:appendix_model_details}

All our experiments, except for~\Cref{sec:efficient_training}, utilize the BART-base model~\cite{DBLP:conf/acl/LewisLGGMLSZ20}\footnote{\url{https://huggingface.co/facebook/bart-base}}. Below, we detail the input format with respect to the K-NRG problem.

\subsection{Formatting the Input}
\label{sec:appendix_input_formatting}

As mentioned in~\Cref{sec:exp-kgnrg}, our input comprises a knowledge snippet $k$ and the dialog history $h$. Here, dialog history is the last five turns in the dialog, with respect to the response. To prepare the input, we assign a fixed number of tokens for each section in the input. We call each section  a bucket. If the actual number of tokens of an input section is less than the total tokens assigned for that bucket, we pad the input to infill the empty tokens. In particular, we provide $32$ tokens for the knowledge snippet $k$ and $25$ tokens for each turn in the dialog history. 

We start the input sequence with the special token $\langle s \rangle$, followed by the knowledge snippet's bucket. Next, we include the dialog history, whose turns use alternate start symbols: $\langle speaker1 \rangle, \langle speaker2 \rangle$. Overall, our input comprises $163$ tokens, 33 knowledge tokens plus 26 turn tokens for each of the 5 turns. On the decoder side, for teacher-forcing, we provide the human response as the input, along with the start token $\langle s \rangle$.

\subsection{Training Details and Hyper-Parameters}
\label{sec:appendix_training}

For training the models, we follow the simple maximum likelihood-based training using ground-truth human responses. It should be re-emphasized that we do not use any of the control knobs during training. Thus, for fine-tuning the BART model on the Topical-Chat data, we train the model for a maximum of 10 epochs with early stopping (patience = 1). The early stopping metric is applied on the average perplexity of the validation set (frequent split). We train with a batch size of 5, gradient accumulation of 4, and learning rate of $6.25e-5$. 

For inference, we follow~\cite{DBLP:conf/inlg/HedayatniaGKLEH20} and utilize nucleus sampling~\cite{DBLP:conf/iclr/HoltzmanBDFC20} with a top-p value of $0.9$. Top-k is set to $0$ and temperature is set to $0.7$. The maximum length of the responses is set to $40$ tokens. We experiment with other values of top-p, but do not observe significant changes in control.

\section{Informativeness Experiments}
\label{sec:appendix_informativeness_exps}

\subsection{Additional Metrics}
\label{sec:appendix_informativeness_additional_metrics}

In this section, we present the extended results with respect to~\Cref{tab:informative_responses_test}. First, we detail the automatic metrics that we consider for the informativeness experiments.

\paragraph{Comparing with Human Responses.} We test the quality of the responses by calculating automatic metrics with respect to the ground-truth human responses. The set of metrics include, perplexity (PPL$_r$), Unigram F1 (F1$_{r}$), BLEU$_{r}$, ROUGE$_{r}$, METEOR$_{r}$, and also the model-based BertScore~\cite{DBLP:conf/iclr/ZhangKWWA20} (B-Score$_{r}$).

\paragraph{Comparing with Knowledge Snippet.} To compare the amount of knowledge incorporated into the response, we calculate the above metrics with the knowledge snippet as the reference. We call these metrics, F1$_k$, BLEU$_k$, ROUGE$_k$, and METEOR$_k$\footnote{In both the settings, we use BLEU-4 and ROUGE-L as the respective metrics.}.

\Cref{tab:appendix_informative_responses_test_freq} and \Cref{tab:appendix_informative_responses_test_rare} present the overall results for automatic metrics across both frequent and rare test sets, respectively. The results in these additional metrics follow similar trends to the metrics discussed in~\Cref{sec:informativeness_results}.

\begin{table*}[t]
    \centering
    \resizebox{0.8\linewidth}{!}{
    \begin{tabular}{|l|c|c|c|c|c|c|}
        \hline
        \multirow{2}{*}{\textbf{Base Model vs. }} & \multicolumn{2}{c|}{\textbf{Fluency}} & \multicolumn{2}{c|}{\textbf{Relevance}} & \multicolumn{2}{c|}{\textbf{Informativeness}} \\
        & p-value & SSD & p-value & SSD & p-value & SSD \\
        \hline
        Knob A (\textit{Knowledge}) & 0.90 & \cellcolor{Red1} No & 0.79 & \cellcolor{Red1} No & 0.001  & \cellcolor{Green1} Yes \\
        Knob A (\textit{Gradual Knowledge}) & 0.90 & \cellcolor{Red1} No & 0.90 & \cellcolor{Red1} No & 0.001 & \cellcolor{Green1} Yes \\
        Knob B ($\boldsymbol{\alpha} = [0.5, 0.5]$) & 0.9 & \cellcolor{Red1} No & 0.13 & \cellcolor{Red1} No & 0.001 & \cellcolor{Green1} Yes \\
        Knob A$+$B (\textit{Gradual Knowledge}, $\boldsymbol{\alpha} = [0.5, 0.5]$) & 0.32 & \cellcolor{Red1} No & 0.052 & \cellcolor{Red1} No & 0.001 & \cellcolor{Green1} Yes \\
        \hline
    \end{tabular}
    }
    \caption{Comparing variants to the base model for statistically significant mean difference in human evaluation scores as per Tukey's HSD test. SSD refers to Statistically Significant Difference between the models for $p<0.001$.}
    \label{tab:appendix_informativess_human_eval_Tukey}
\end{table*}

\subsection{Human Evaluation Details}
\label{sec:appendix_informativeness_human_eval}

For the human evaluation of informativeness, relevance and fluency, we utilize the questionnaire discussed in~\Cref{sec:informativeness_evaluation_setup}. We randomly sample 200 dialog instances from the combined test sets of frequent and rare splits in Topical-Chat (100 each). Each instance has the dialog history ($h$) with five dialog turns and the provided knowledge snippet ($k$). However, we notice that the top-selected knowledge snippet for a particular dialog context may not always be entirely relevant for the response. This would affect the human evaluations as we specifically ask the annotators to prefer responses where facts from the knowledge snippet is manifested in generated response. Thus, we first filter the test sets before sampling the 200 instances. Specifically, we calculate the ROUGE metric between the knowledge snippet and the human response, and only consider the set of dialog contexts that have a higher value than the mean ROUGE value of 0.2\footnote{Mean ROUGE between knowledge snippet and human response over the Topical-Chat training set is 0.2.}. This filtration ensures that the knowledge snippet is relevant to the dialog context and thus can be a good test bed for measuring control over informativeness.

We use Amazon Mechanical Turk as the annotation platform and appoint three annotators per response sample across all model variants. To ensure high quality for annotations, we opt for annotators that are familiar with dialog evaluation and have a high overall performance as Turkers ($95\%$ or higher approval rate and more than $5000$ approved HITs). 

\paragraph{Results.} The main results of the average Likert-based scores are summarized in~\Cref{tab:informative_responses_test}. For fluency, relevance, and informativeness, the respective inter-annotator agreement (IAA) using Krippendorff's alpha are as follows: $0.545$, $0.354$, $0.373$. As relevance and informativeness are scored on a wider scale of 1 to 5, we categorize this 5-scale Likert scale into three bins comprising the values [1,2], [3], and [4,5]. As seen in the IAA values, we achieve high agreements for fluency. For relevance and informativeness, our IAA scores are similar to \cite{DBLP:conf/inlg/HedayatniaGKLEH20} where the annotations were on a ranking-based format and not Likert-based. It is known in the literature that Likert-based annotations, due to factors like personal bias of annotators, are prone to have lower IAA scores~\cite{DBLP:conf/inlg/LeeGMWK19}. Having said that, we choose this process as it provides a descent average value of each model variant~\cite{DBLP:journals/corr/abs-2101-06561}. Comparing the mean statistics of the variants, we perform statistical significance tests between all the variant pairs using the Tukey's HSD test. We find that compared to the base model (no control knobs applied), none of the controlled variants have fluency or relevance scores that are statistically significant in difference. In contrast, all the variants achieve statistically significantly higher informativeness scores. This highlights that the variants are able to improve on informativeness without compromising on fluency and relevance.

\section{Context Augmentation Experiments}
\label{sec:appendix_control_exps}

\subsection{Human Evaluation Details}
\label{sec:appendix_control_human_eval}

Similar to \Cref{tab:appendix_informativess_human_eval_Tukey}, we perform statistical tests for the variants introduced in~\Cref{sec:control_experiments}. The results are summarized in~\Cref{tab:appendix_control_human_eval_Tukey}. As seen in the Table, we do not find any statistically significant difference between the controlled variants when compared to the base model in both fluency and relevance.

\begin{table}[t]
    \centering
    \resizebox{0.95\linewidth}{!}{
    \begin{tabular}{|l|c|c|c|c|}
        \hline
        \multirow{2}{*}{\textbf{Base Model vs. }} & \multicolumn{2}{c|}{\textbf{Fluency}} & \multicolumn{2}{c|}{\textbf{Relevance}} \\
        & p-value & SSD & p-value & SSD \\
        \hline
        C & 0.623 & \cellcolor{Red1} No & 0.802 & \cellcolor{Red1} No \\
        C$+$A & 0.871 & \cellcolor{Red1} No & 0.263 & \cellcolor{Red1} No  \\
        C$+$B & 0.900 & \cellcolor{Red1} No & 0.900 & \cellcolor{Red1} No  \\
        C$+$A$+$B & 0.900 & \cellcolor{Red1} No & 0.199 & \cellcolor{Red1} No  \\
        \hline
    \end{tabular}
    }
    \caption{Comparing variants to the base model for statistically significant mean difference in human evaluation scores. SSD = ``Yes" means Statistically Significant Difference between the models for $p<0.001$.}
    \label{tab:appendix_control_human_eval_Tukey}
\end{table}

\begin{table}[t!]
\centering\renewcommand\cellalign{lc}
\setcellgapes{3pt}\makegapedcells
\resizebox{\linewidth}{!}{
    \begin{tabular}{|l|l|l|} \hline
        \textbf{PropQ} & \textbf{SetQ} & \textbf{ChoiceQ} \\
        \hline
        Do you like & How are you & Or are you \\
        Do you know & How much do you & Or do you \\
        Do you watch & How can you & Is it just \\
        Do you have & What do you & Is there a reason \\
        Have you ever & What kind of & Do you think or \\
        Are you a & What did you think & \\
        & Why is that & \\
        & Why do you & \\
        \hline
    \end{tabular}
}
\caption{ \footnotesize Control phrases for fine-grained question types.}
\label{tbl:biasing_phrases_example}
\end{table}

\subsection{Control Phrases}
\label{sec:biasing_phrases_example}

\Cref{tbl:biasing_phrases_example} presents the control phrases that we use for the respective fine-grained question generation.

\subsection{Human Evaluation for Control Classifier}
\label{sec:appendix_control_classifier_human_eval}

For investigating the reliability of the RNN-based control classifier, we proceed to check its accuracy with respect to human ground truths. We start by sampling $300$ sentences from the test set and ask two human annotators to annotate each sentence with the reduced tag-set: \textit{PropQ}, \textit{ChoiceQ}, \textit{SetQ}, \textit{Feedback}, \textit{Salutation}, \textit{Statement}, and \textit{Others}. Here, the category \textit{Others} collate infrequent dialog acts, such as \textit{Directives}\footnote{Full tag-set is available in~\cite{DBLP:conf/inlg/HedayatniaGKLEH20}}.

The annotators get a very high inter-annotator agreement with Krippendorff's alpha $0.8$. For the conflicts, we employ a third annotator to break the ties. With this, we get the ground truth annotations over the $300$ sentences.~\Cref{tbl:control_classifier_examples} demonstrates some of the sentences with the human annotation.

\begin{table}[t!]
\centering\renewcommand\cellalign{lc}
\setcellgapes{3pt}\makegapedcells
\resizebox{\linewidth}{!}{
    \begin{tabular}{|c|l|} \hline
        \textbf{Sentence} & \textbf{Dialog Act} \\
        \hline
        \makecell{Do you know what the University of \\ Iowa's locker room is?} & \textit{PropQ} \\
        \makecell{What about you?} & \textit{SetQ} \\
        \makecell{Or does it become a problem?} & \textit{ChoiceQ} \\
        \makecell{I haven't seen that one, but I have heard \\  that he tried to retire the first time.} & \textit{Statement} \\
        \makecell{Wow that is a lot.} & \textit{Feedback} \\
        \makecell{I hope you have a good day too!} & \textit{Salutation} \\
        \hline
    \end{tabular}
}
\caption{ \footnotesize Samples of sentences from Topical-Chat annotated with dialog acts.}
\label{tbl:control_classifier_examples}
\end{table}

Next, we automatically annotate the sentences with both the off-the-shelf SVM~\cite{DBLP:conf/coling/MezzaCSTR18} and our RNN-based taggers. \Cref{tbl:control_classifier_f_score} shows the classification results, where the RNN-based classifier achieves a higher F1-score ($0.84$) than the SVM ($0.77$). Notably, the F1 (and particularly the precision) of the question categories are very high which establishes this classifier as a reliable control evaluator.

\begin{table}[t!]
\centering\renewcommand\cellalign{lc}
\setcellgapes{3pt}\makegapedcells
\resizebox{\linewidth}{!}{
    \begin{tabular}{|c|ccc|ccc|} \hline
        \multirow{2}{*}{\textbf{Dialog Acts}} & \multicolumn{3}{c|}{SVM}  & \multicolumn{3}{c|}{RNN}   \\
        & \textbf{Precision} & \textbf{Recall} & \textbf{F1} & \textbf{Precision} & \textbf{Recall} & \textbf{F1} \\
        \hline
        PropQ & 1.00 & 0.84 & 0.91 & 0.98 & 1.00 & \textbf{0.99}\\
        SetQ & 0.88 & 0.80  & 0.83 & 1.00 & 0.91 & \textbf{0.95}\\
        ChoiceQ & 1.00  & 1.00  & \textbf{1.00} & 1.00 & 0.82 & 0.90 \\
        Statement & 0.60 & 0.63 & \textbf{0.62} &  0.53 & 0.76 & \textbf{0.62} \\
        Feedback & 0.90 & 0.52 & 0.66 & 0.81 & 0.69 & \textbf{0.75} \\
        Salutation & 0.86 &  0.93  & 0.89 & 0.93 & 0.96 & \textbf{0.95} \\
        Other & 0.00 &  0.00 & 0.00 & 0.00 & 0.00 & 0.00 \\
        \hline
        Accuracy & - & - & 0.72 & - & - & \textbf{0.83} \\
        Weighted Avg. & 0.85 & 0.72 & 0.77 & 0.85 & 0.83 & \textbf{0.84} \\
        \hline
    \end{tabular}
}
\caption{ \footnotesize F1 score of SVM and RNN model predictions over human annotated ground truth dialog acts. Numbers in boldface represent the higher F1-score between SVM and RNN models.}
\label{tbl:control_classifier_f_score}
\end{table}

\subsection{Human Evaluation for Detecting Questions Based on  `?'}
\label{sec:appendix_question_detector_human_pref}

We measure the reliability of using `?' as a proxy for valid questions. For this, we perform a human evaluation similar to~\Cref{sec:appendix_control_classifier_human_eval}, where we ask two annotators to identify questions. In particular, we take the $200$ generations from the C$+$A$+$B model in~\Cref{tab:question_control} and ask the annotators to mark a response as `yes' if it contains a valid question and `no' otherwise. As this is an objective question, we find only four disagreements between the annotators, which we resolve after discussing them. This provides the ground-truth question annotation. We compare this ground truth with the proxy-based approach that we employ, and present the results in~\Cref{tbl:question_classifier_f_score}. The numbers show very high similarity between `?' based marker and human annotations. Particularly, the `?' based marker obtains a 0.99 precision which demonstrates that the marker does not falsely count questions as positive. For recall, it has 0.87 which upon investigations we notice that they are all of the type ``I wonder" where there is no explicit question mark. An example of such a question is: ``\textit{I wonder if it was the NFL's tracking chips or the tracking chips embedded in players shoulder pads to record a players every move on the field.}". Additionally, the Krippendorff's alpha between the `?' based and human annotations are 0.935. These results show that, in general, the `?' based counts that are shown in~\Cref{tab:question_control} and related tables are highly reliable.

\begin{table}[t!]
\centering\renewcommand\cellalign{lc}
\setcellgapes{3pt}\makegapedcells
\resizebox{0.8\linewidth}{!}{
    \begin{tabular}{|c|ccc|} \hline
        \multirow{2}{*}{\textbf{isQuestion?}} & \multicolumn{3}{c|}{`?' marker for Question}   \\
        & \textbf{Precision} & \textbf{Recall} & \textbf{F1}  \\
        \hline
        yes & 0.99 & 0.87 & 0.92 \\
        no & 0.90 & 0.99  & 0.94 \\
        \hline
        Accuracy & - & - & 0.94 \\
        Weighted Avg. & 0.94 & 0.94 & 0.93 \\
        \hline
    \end{tabular}
}
\caption{ \footnotesize Comparison of proxy-based vs. human-based question detection.}
\label{tbl:question_classifier_f_score}
\end{table}

\subsection{Qualitative Examples}
\label{sec:appendix_qualitative_examples}

\Cref{tab:appendix_question_control_examples} presents a few dialog instances from the test set that we randomly sample. We show the generations by the variants detailed in \Cref{sec:control_experiment_setup}.

\Cref{tab:appendix_fine-grained-question-examples} presents examples for the fine-grained question types, and feedback and sentiment attributes. The generations are by the C$+$A$+$B model. While we observe the attributes in the generated samples, for the positive sentiment in the second example, we notice that the model does not generate with positive sentiment when the context is not appropriate for it. This is an important skill in grounded tasks like dialog response generation.

\begin{table*}[h]
\centering\renewcommand\cellalign{lc}
\setcellgapes{3pt}\makegapedcells
    \begin{subtable}[h]{0.49\textwidth}
        \resizebox{\linewidth}{!}{
        \begin{tabular}{|l|l|} \hline
            \multirow{ 5}{*}{\makecell{\textbf{Previous} \\ \textbf{Turns}}} & \textbf{A}: hi how are you? do you play video games?\\
            & \makecell{\textbf{B}: I'm doing well how are you? Yes I like playing \\ video games, especially on Xbox!}\\
            & \makecell{\textbf{A}: I have not played Xbox or PS.\\ whhich do you prefer? why?}\\
            & \makecell{\textbf{B}: I like xbox more because honestly \\ it's the only system I've had. \\ I remember getting one way back in 2001.}\\
            & \makecell{\textbf{A}: Microsoft gets many things wrong, but \\ ive heard good things about xbox games. \\ i played games as a kid, but now, im not interested. \\ What games do you like?}\\
            \hline
            \makecell{\textbf{Knowledge}} & \makecell{\itshape ``\itshape there are controllers created for quadriplegics\\ \itshape so they can play XBOX 360 and other consoles.''} \\ 
            \hline
            \textbf{Response} & \\
            \colorbox{Gray1}{Base Model} & \makecell{\textbf{B}: I agree with you. I've heard that they \\ created controllers for quadriplegics so they can play \\ xbox 360 and other consoles.}\\
            \colorbox{Blue1}{C} & \makecell{\textbf{B}: I agree with you on that one! \\ I'm not surprised that they created controllers for quadriplegics, \\ I didn't know that. }\\
            \colorbox{Blue1}{C$+$A} & \makecell{\textbf{B}: \colorbox{Green1}{Did you know there are controllers created for} \\ \colorbox{Green1}{quadriplegics so they can play xbox 360 and other consoles?}}\\
            \colorbox{Blue1}{C$+$B} & \makecell{\textbf{B}: \colorbox{Green1}{Did you know that there} \\ \colorbox{Green1}{are controllers created for quadriplegics so they can play xbox 360} \\ \colorbox{Green1}{and other consoles?} That's really cool!}\\
            \colorbox{Blue1}{C+A+B} & \makecell{\textbf{B}: I agree with that! There are controllers created \\ for quadriplegics so they can play \\ XBOX 360 and other consoles. \\ \colorbox{Green1}{I wonder how many of those are on the first one?}}\\
            \hline
        \end{tabular}
        }
    \end{subtable}
    \hfill
    \begin{subtable}[h]{0.45\textwidth}
        \resizebox{\linewidth}{!}{
        \begin{tabular}{|l|l|} \hline
            \multirow{ 5}{*}{\makecell{\textbf{Previous} \\ \textbf{Turns}}} & \makecell{\textbf{A}: Good morning!  Have you heard of Stan Lee?  \\ Did you know that he did cameos in \\ all of the Marvel movies?}\\
            & \makecell{\textbf{B}: I have! He was such a good writer! \\ I love looking for him in his films!}\\
            & \makecell{\textbf{A}: I do too and I agree, he was a good writer.  \\Sometimes he even speaks, like in Captain America,\\ the Winter Soldier.}\\
            & \makecell{\textbf{B}: I didn't know that! What else was he in?}\\
            & \makecell{\textbf{A}: My favorites are the Iron Man appearances. \\ It seems like he had fun with those because \\ in one Tony Stark thinks he's Hugh Hefner\\ and in another he thinks he's Larry King!}\\
            \hline
            \makecell{\textbf{Knowledge}} & \makecell{\itshape ``\itshape Luckily, the titular character \\ \itshape prevents him from getting hit by a bus.''} \\ 
            \hline
            \textbf{Response} & \\
            \colorbox{Gray1}{Base Model} & \makecell{\textbf{B}: That's hilarious! I love that character! \\ I loved him in The Iron Man!}\\
            \colorbox{Blue1}{C} & \makecell{\textbf{B}: That's a good one! \\ \colorbox{Green1}{Did you know he was the titular character?}}\\
            \colorbox{Blue1}{C$+$A} & \makecell{\textbf{B}: I never knew that. \colorbox{Green1}{I wonder if it is the same} \\ \colorbox{Green1}{character that killed off the Hulk? }}\\
            \colorbox{Blue1}{C$+$B} & \makecell{\textbf{B}: That's a lot of fun! I just can't \\ believe he was actually doing that!}\\
            \colorbox{Blue1}{C+A+B} & \makecell{\textbf{B}: I love the first few Iron Man appearances, \\ but the last one was a little odd. \\ \colorbox{Green1}{I wonder if he ever got hit by a bus?}}\\
            \hline
        \end{tabular}
        }
    \end{subtable}
    \caption{ \footnotesize Randomly sampled dialogs along with the generations by control knobs from  \Cref{sec:control_experiment_setup}. Questions in the turns are highlighted in \colorbox{Green1}{green}.}
    \label{tab:appendix_question_control_examples}
\end{table*}

\begin{table*}[h]
\centering\renewcommand\cellalign{lc}
\setcellgapes{3pt}\makegapedcells
    \begin{subtable}[h]{0.49\textwidth}
        \resizebox{\linewidth}{!}{
        \begin{tabular}{|l|l|} \hline
            \multirow{ 5}{*}{\makecell{\textbf{Previous} \\ \textbf{Turns}}} & \makecell{\textbf{A}: Did you know that the University of Iowa's \\ locker room is painted pink? I wonder why?}\\
            & \makecell{\textbf{B}: I think I did hear something about that.  \\I imagine it is an attempt to psych the other team out.}\\
            & \makecell{\textbf{A}: So, it would be in the visiting team's locker \\ room but not their own?}\\
            & \makecell{\textbf{B}: Right. Teams do all kinds of things to bother \\ the competition. I've heard of teams having heated benches \\ in the winter for themselves but not for the visitors.}\\
            & \makecell{\textbf{A}: I would hate a cold bench. Then again, \\ I wouldn't want to be some place that \\ cold or watching football.}\\
            \hline
            \makecell{\textbf{Knowledge}} & \makecell{\itshape ``That the highest score ever in a  \\
            \itshape football game occurred in 1916 when Georgia Tech  \\ 
            \itshape defeated Cumberland 222-0''} \\ 
            \hline
            \textbf{Response} & \\
            \colorbox{Gray1}{PropQ} & \makecell{\textbf{B}: \colorbox{Gray1}{Do you know what the highest score ever} \\ \colorbox{Gray1}{was in a football game?}}\\
            \colorbox{Blue1}{SetQ} & \makecell{\textbf{B}: \colorbox{Blue1}{What kind of football do you think they play?} \\ I bet they score like 222 points}\\
            \colorbox{almond}{Feedback} & \makecell{\textbf{B}: \colorbox{almond}{Yeah, that's pretty extreme.} \\ I'd rather not go to Georgia Tech. \\ They beat Cumberland 222-0. \\ That's pretty crazy.} \\
            \colorbox{Green1}{Positive} & \makecell{\textbf{B}: That's right. Georgia Tech beat Cumberland 222-0 in 1916. \\ \colorbox{Green1}{It was a good game}. \\ They must have been exhausted.} \\
            \hline
        \end{tabular}
        }
    \end{subtable}
    \hfill
    \begin{subtable}[h]{0.5\textwidth}
        \resizebox{\linewidth}{!}{
        \begin{tabular}{|l|l|} \hline
            \multirow{ 5}{*}{\makecell{\textbf{Previous} \\ \textbf{Turns}}} & \textbf{A}: Hi, do you watch nfl?\\
            & \makecell{\textbf{B}: Yea, it's crazy the Seahawks released Richard Sherman}\\
            & \makecell{\textbf{A}: yeah that is. did you know women can play too}\\
            & \makecell{\textbf{B}: I've heard there is no official rule barring female \\ players in the NFL, I wonder if there will ever be one}\\
            & \makecell{\textbf{A}: yeah me too. did you hear about that \\ rule for new orleans cheerleaders?}\\
            \hline
            \makecell{\textbf{Knowledge}} & \makecell{\itshape ``New Orleans Saints cheerleaders are  \\
            \itshape forbidden from eating in the same restaurant as any NFL player  \\ 
            \itshape and if they are already dining at a restaurant and an NFL player  \\
            \itshape comes in after, the cheerleaders are required to leave.''} \\ 
            \hline
            \textbf{Response} & \\
            \colorbox{Gray1}{PropQ} & \makecell{\textbf{B}:  \colorbox{Gray1}{Do you think they can get away with it?} \\ Or is it just a rule? I mean, if \\they're eating at the same restaurant as the players, then it's okay. }\\
            \colorbox{Blue1}{SetQ} & \makecell{\textbf{B}: \colorbox{Blue1}{What is the rule?} It seems like \\a lot of sexist stuff to me. I'd be happy \\if the cheerleaders were allowed to eat at the \\ same restaurant as the players }\\
            \colorbox{almond}{Feedback} & \makecell{\textbf{B}: \colorbox{almond}{That's pretty sexist.} \\ I wonder if they'll ever get a job in the NFL \\ and get fired for being sexist and homophobic} \\
            \colorbox{Green1}{Positive} & \makecell{\textbf{B}: That's crazy, I wonder how they treat \\ the cheerleaders in the NFL} \\
            \hline
        \end{tabular}
        }
    \end{subtable}
    \caption{ \footnotesize Randomly sampled dialogs controlled by different fine-grained attributes (\Cref{tab:fine-grained_questions}). Interestingly, the model does not generate a positive response when it is not appropriate to the context.}
    \label{tab:appendix_fine-grained-question-examples}
\end{table*}

\end{document}